\pdfoutput=1
\documentclass[10pt,twocolumn,letterpaper]{article}

\usepackage[pagenumbers]{cvpr}

\usepackage{cuted} %

\usepackage{currfile} %

\usepackage{caption} %

\usepackage{multirow} %
\usepackage[normalem]{ulem} %
\definecolor{BestBg}{RGB}{255,185,185}
\definecolor{SecondBg}{RGB}{255,238,150}
\newcommand{\best}[1]{{\setlength{\fboxsep}{1pt}\hspace{-1pt}\colorbox{BestBg}{#1}\hspace{-1pt}}} %
\newcommand{\second}[1]{{\setlength{\fboxsep}{1pt}\hspace{-1pt}\colorbox{SecondBg}{#1}\hspace{-1pt}}}
\definecolor{EditAdd}{rgb}{0.00,0.35,0.85}
\definecolor{EditDel}{rgb}{0.85,0.10,0.10}
\definecolor{EditNote}{rgb}{0.90,0.45,0.00}
\newif\ifshowedits
\showeditsfalse

\definecolor{EditPur}{rgb}{0.55,0.10,0.70}

\usepackage{enumitem}
\makeatletter
\renewcommand\paragraph{\@startsection{paragraph}{4}{\z@}%
  {0.5ex}{-0.8em}{\normalfont\normalsize\bfseries}}
\makeatother
\setlist{nosep,leftmargin=*}

\makeatletter
\def\abstract{%
   \iftoggle{cvprpagenumbers}{}{\thispagestyle{empty}}%
   \centerline{\large\bf Abstract}%
   \vspace*{4pt}\noindent%
   \it\ignorespaces%
}
\makeatother

\makeatletter
\renewcommand\paragraph{\@startsection{paragraph}{4}{\z@}{1.2ex \@plus .3ex \@minus .2ex}{-1em}{\normalfont\normalsize\bfseries}}
\makeatother
\usepackage{array}

\definecolor{cvprblue}{rgb}{0.21,0.49,0.74}
\usepackage[pagebackref,breaklinks,colorlinks,allcolors=cvprblue]{hyperref}

\def\confName{CVPR}
\def\confYear{2027}

\title{Layered Director Notes for Facial Motion Generation}

\author{
Shuran Zhao$^{1}$\quad Tian Wang$^{1}$\quad Daohan Lu$^{2}$\quad Guanzhou Ji$^{1}$\\
$^{1}$Northeastern University\qquad $^{2}$New York University
}

\begin{document}
\maketitle
\begin{strip}
\centering
\includegraphics[width=0.85\linewidth]{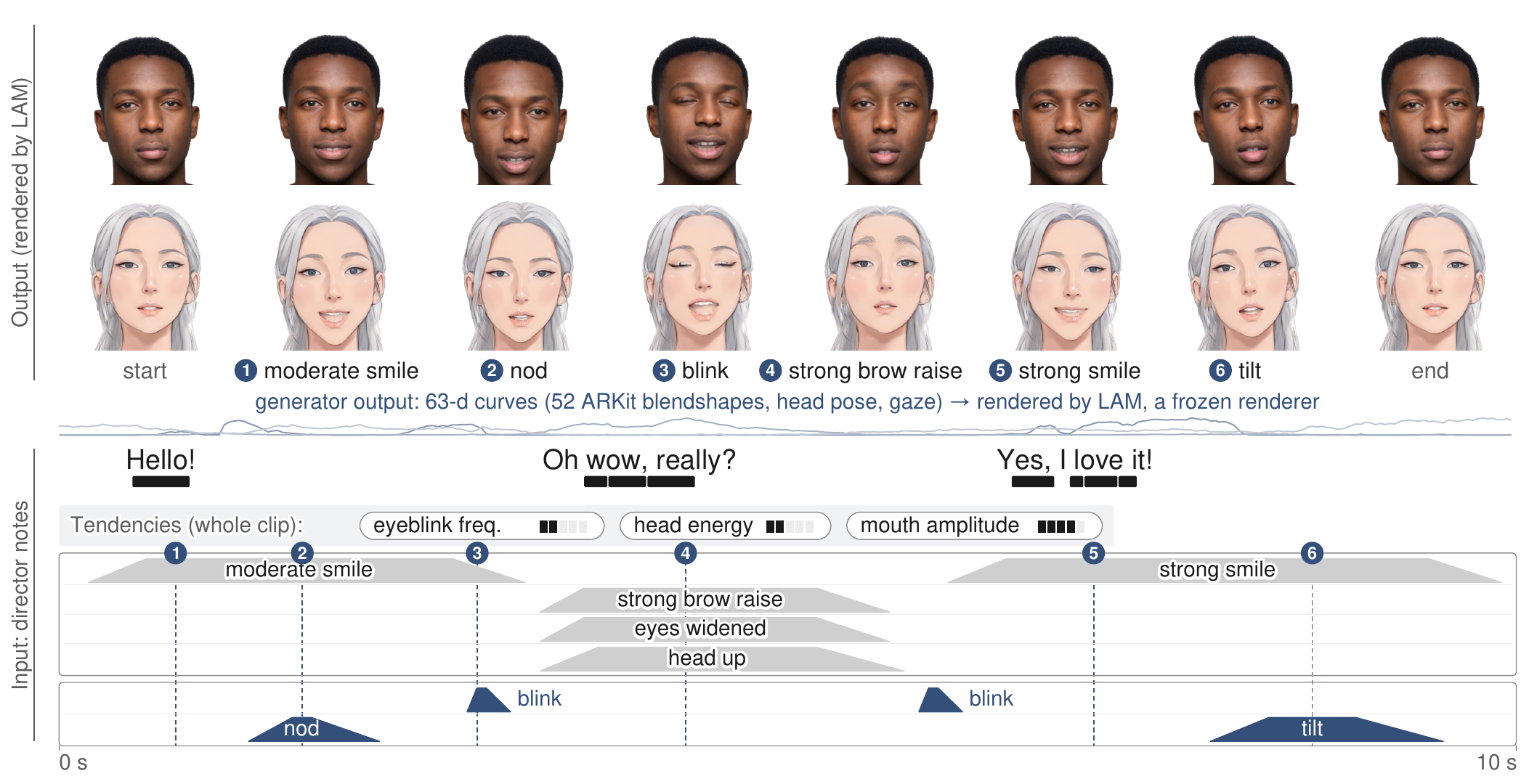}
\captionsetup{hypcap=false}
\captionof{figure}{\textbf{Layered Director Notes (LDN): a new control signal for facial motion generation.} A deterministic encoder converts any tracked face video into such notes without manual labels, and a diffusion model, trained from random initialization, generates a complete performance (blendshape weights, head pose and gaze) from sparse notes and the spoken lines. Unlike audio, text or single-layer timelines, the notes specify how strong a motion is and how it begins and ends, at three temporal scales: tendencies (top box), states (gray) and events (blue). The same notes drive different faces, rendered with the frozen LAM~\cite{he2025lam} from one portrait each, generated with Gemini 3 Pro Image; mouth motion follows the script timing, not audio.}
\captionsetup{hypcap=true}
\label{fig:teaser}
\end{strip}

\begin{abstract}
Directing a talking character's face requires controlling how strong each motion is, how it begins and ends, and how motions overlap, which existing control signals only partly provide. We present Layered Director Notes (LDN), a sparse language describing a performance as overall tendencies, seconds-long states and momentary events, each with an intensity and onset and offset timing. A diffusion model, trained on randomly masked notes, generates the full performance (blendshapes, head pose and gaze) from the notes and a timed script, executing specified events and completing unspecified spans naturally. A deterministic encoder extracts LDN from tracked video without manual labels, providing training data and automatic evaluation. Language models author LDN from a few examples without fine-tuning on LDN data. LDN controls intensity more precisely than single-layer timelines, and users prefer our motion over an audio-driven baseline and do not distinguish it from real motion. The code and checkpoints are available \href{https://github.com/shurandaa/Layered-Director-Notes-for-Facial-Motion-Generation}{here}.
\end{abstract}

\section{Introduction}
\label{sec:intro}

Facial motion generation has been widely studied for digital humans, virtual assistants and social robots. As language models increasingly drive verbal interactions between users and virtual characters, a natural next step is to let them understand and direct a character's facial performance while it talks. Existing methods control facial motion with four kinds of signals: audio-driven methods generate motion from speech~\cite{fan2022faceformer,xing2023codetalker,danecek2023emote,peng2023emotalk,sun2024diffposetalk,aneja2024facetalk,zhao2024media2face,xu2024vasa,nvidia2025audio2face}, text-guided methods describe the motion in words, on top of speech~\cite{zhong2024expclip,wang2025instructavatar} or alone~\cite{xu2025whenwordssmile}, sometimes as a timed plan derived from audio~\cite{huang2025think2sing,lyu2026auhead}, keyframe- and mask-based methods specify the motion parameters directly at selected frames~\cite{thambiraja2023diface,park2023said}, as in body motion~\cite{cohan2024condmdi,xie2024omnicontrol}, and timeline methods place labeled actions in time~\cite{ma2025timeline}, as in body motion~\cite{petrovich2024stmc}. However, directing facial performance with language models remains largely unexplored~\cite{ma2025timeline,mahadevan2024genem,abbo2026expressive,durupinar2025prompt,wang2023agentavatar}.

Our goal is to generate a complete facial performance from sparse director notes written in a structured language, so that a language model or a user can edit facial expressions, gaze and 3D head movement by changing the notes. We do not condition on audio: the notes are authored and revised before any speech is synthesized, and a model driven by the voice would infer expressions and head motion from it, competing with the notes; where audio exists, it can drive the lips. This requires a representation that captures overlapping motions at different time scales, together with the intensity of each motion and how it starts and ends. When a time span is left unspecified, the model should still perform it naturally rather than hold it neutral. Existing families provide some of these abilities, none all (\cref{sec:related}).

We present Layered Director Notes (LDN), a sparse, layered language for directing facial performance, and the LDN system built around it. The LDN system consists of three parts: the notes themselves, a deterministic encoder and decoder that translate between motion curves, tracked from video by the pretrained and frozen MediaPipe~\cite{lugaresi2019mediapipe}, and notes, and a generator, a diffusion model that we train from random initialization, that produces the whole performance from sparse notes and the script. The notes describe a performance in three layers by time scale: (1) tendencies for the whole performance, such as how often the character blinks; (2) states that last for seconds, such as a moderate smile; and (3) events at single moments, such as a nod or a blink. Each note carries an intensity level, an optional value and four time points (\cref{sec:language}). The notes are sparse, and our generator fills in everything unspecified with natural motion that follows the script (the lines to be spoken, with their word timing) and the specified tendencies, instead of treating it as neutral~\cite{ma2025timeline} or completing it from sparse keyframes~\cite{wu2025keyframeface} (\cref{sec:generator}). The encoder and decoder are rule-based and training-free, with constants calibrated once on the training set, and require no manual annotation (\cref{sec:codec}). Our contributions are:
Intensity, phases and layers exist individually in earlier formats~\cite{kopp2006bml,ma2025timeline}; we make them one system that is extracted from video, completed where unspecified and checked by re-encoding.
\begin{itemize}[leftmargin=*]
\item \textbf{Layered director notes} (\cref{sec:language}), a sparse language that language models author without fine-tuning (\cref{sec:exp_llm}) and that gives finer control than a single-layer timeline, from the intensity of each action and how it starts and ends to the style of the whole performance (\cref{sec:exp_timeline}).
\item \textbf{A deterministic encoder and decoder} (\cref{sec:codec}) that translate the tracked motion of any video into LDN and back, with no trained network and no manual labels; fitted once on one dataset, they transfer to unseen datasets and agree with a human annotator on blinks and head and gaze timing (\cref{sec:exp_codec}).
\item \textbf{A sparse-notes generator}, trained from random initialization, that keeps every specified event within a per-channel tolerance of the decoded motion and completes everything unspecified with motion much more natural than holding it neutral, interpolating or retrieving clips (\cref{sec:exp_main}); in a user study, participants preferred it over retrieval and an audio-driven baseline and showed no significant preference between it and real motion (\cref{sec:exp_user}).
\item \textbf{A cycle-consistency evaluation} (\cref{sec:exp_setup}) in which the same encoder reads the generated motion back into LDN, making the evaluation automatic and reproducible; on rendered videos, the tracker followed by our encoder recovers over three times as many specified events as a vision--language model (\cref{sec:exp_vlm}).
\end{itemize}
\noindent Formulas, implementation details and further results are in the supplementary material.

\section{Related Work}
\label{sec:related}

\begin{figure*}[t]
\centering
\includegraphics[width=\textwidth]{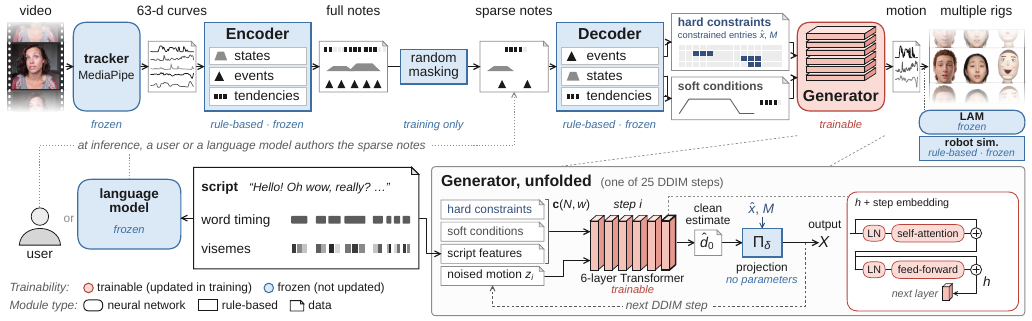}
\caption{\textbf{Overview of Our Pipeline.} The encoder extracts the full director notes from the tracked video; random masking constructs the sparse training input. The decoder converts specified events into constrained entries ($\hat x$, $M$) and states and tendencies into soft conditions. A Transformer predicts the clean motion, which is projected onto the constrained entries at every DDIM step ($\Pi_\delta$).}
\label{fig:overview}
\end{figure*}

\noindent\textbf{Speech-driven facial animation.} These methods map audio to natural facial motion~\cite{fan2022faceformer,xing2023codetalker,richard2021meshtalk,peng2023selftalk}, with emotion or style control~\cite{danecek2023emote,peng2023emotalk,sun2024diffposetalk}, diffusion~\cite{aneja2024facetalk,park2023said,thambiraja2023diface}, head pose~\cite{chen2024diffsheg,xu2024vasa,zhao2024media2face} and industrial tools~\cite{nvidia2025audio2face}; keyframe editing~\cite{park2023said,thambiraja2023diface} and text prompts~\cite{zhao2024media2face} only adjust the audio-driven performance, whereas the notes are authored before any audio exists. Cafe-Talk adds fine-grained action-unit control with adjustable intensity, but still requires speech~\cite{chen2025cafetalk}. Our generator requires no audio and takes the timed script instead, as PhonemeNet does for lip motion~\cite{witzig2025phonemenet}; beyond the lips, it generates directed expressions, head pose and gaze.

\noindent\textbf{Text, keyframe and timeline control.} Text-guided methods set a clip's emotion or motion~\cite{zhong2024expclip,wang2025instructavatar}, map a free-text prompt to a sequence of 3D face parameters~\cite{xu2025whenwordssmile,song2026tdmmlm}, or plan timed subtitles or action unit (AU) sequences from audio and lyrics~\cite{huang2025think2sing,lyu2026auhead}. Keyframe- and mask-based methods for faces~\cite{thambiraja2023diface,park2023said} and bodies~\cite{cohan2024condmdi,xie2024omnicontrol,karunratanakul2024dno} follow the given frames closely, but keyframes authored by language models yield far less natural motion (\cref{sec:exp_llm}); Timelines for faces~\cite{ma2025timeline} and bodies~\cite{petrovich2024stmc} are precise in timing, but the annotation of Timeline Control assigns one class per region and frame (eye closure is labeled separately and may overlap other eye motions) and frames with no label mean Neutral~\cite{ma2025timeline}, so apart from blinks its generator never sees overlapping actions within a region, intensity exists only in a few classes such as SoftSmile, and intervals have start and end frames but no onset or offset phase. LDN overlays events on states and gives each note an intensity, an optional value and four time points (\cref{sec:exp_codec}).

\noindent\textbf{Facial motion annotation.} AUs are coded by trained coders~\cite{ekman1978facial}; attribute labels mark the start and end of dynamic attributes~\cite{yu2023celebvtext} or summarize AUs with a language model~\cite{wang2023agentavatar,wu2024mmhead}, and Timeline Control names its clusters by hand~\cite{ma2025timeline}. MotionScript converts body motion into structured text without training~\cite{yazdian2023motionscript}, and TDMM-LM describes 3D face parameters in free text with a language model~\cite{song2026tdmmlm}. None of them provides a deterministic decoder back to motion, whereas the LDN system translates between tracked motion and notes both ways without manual labels (\cref{sec:exp_codec}).

\noindent\textbf{Behavior markup and language models.} BML specifies the timing, amount, onset and offset of behaviors, facial expressions included, but leaves the motion to a separate realizer~\cite{kopp2006bml}. The language-model systems below act at the level of whole gestures or turns: GenEM composes robot skills~\cite{mahadevan2024genem}, Expressive Furhat generates one facial gesture per dialogue turn, and its authors name the missing synchronization with speech as the main limitation~\cite{abbo2026expressive}, and Prompt-to-Animation writes FACS animations per scenario~\cite{durupinar2025prompt}. AgentAvatar plans facial motion as free text, drives it with a fine-tuned language model that outputs motion tokens and scores the result with ChatGPT and a small user study~\cite{wang2023agentavatar}. LDN, by contrast, can be extracted from the tracked motion of any video and follows word timing (\cref{sec:exp_codec}), can be sparse with specified events kept within a tolerance (\cref{sec:exp_main}), gives about a third of the FMD of FACS or blendshape keyframes authored by language models (\cref{sec:exp_llm}), and is checked automatically by re-encoding.

\section{Method}
\label{sec:method}

\Cref{fig:overview} shows the overall workflow. Videos are first tracked into per-frame motion curves (\cref{sec:repr}).
The encoder translates the curves into full director notes, and the decoder translates the notes back into curves
(\cref{sec:codec}). During training, we randomly mask entries of the notes and train the generator to perform the
whole clip from the sparse notes and the script timing (\cref{sec:generator}). At inference, the notes are specified by
an off-the-shelf language model or authored or edited directly by a user; for evaluation, the generated motion is read
back by the same encoder (\cref{sec:eval}). Because the encoder and decoder are rule-based with constants calibrated once,
the notes mean the same thing in training, generation and evaluation, and every specified element can be traced to the
curves it controls.

\subsection{Motion Representation}
\label{sec:repr}

Facial motion is a 30\,Hz sequence $X=(x_1,\dots,x_T)$ of 63-dimensional frames $x_t=[b_t,p_t,r_t,g_t]$: the 52
blendshapes $b_t\in[0,1]^{52}$ estimated by MediaPipe~\cite{lugaresi2019mediapipe} (named as in ARKit~\cite{apple2017arkit}),
the head translation $p_t\in\mathbb R^3$, the head rotation $r_t\in\mathbb R^6$ in the 6D
representation~\cite{zhou2019continuity}, and the gaze $g_t\in\mathbb R^2$ (vertical and horizontal iris offset). To animate a portrait, we use LAM~\cite{he2025lam}, which reconstructs a FLAME-driven~\cite{li2017flame}
Gaussian head from one photo; a ridge regression fitted on tracked training frames maps the blendshapes to FLAME
expression and jaw parameters. The director language expresses head rotation in degrees relative to the reference
orientation $R_\text{ref}$: $\omega_t=\tfrac{180}{\pi}\log(R_\text{ref}^{\top}R(r_t))^{\vee}$.

\begin{figure*}[t]
\centering
\includegraphics[width=0.85\textwidth]{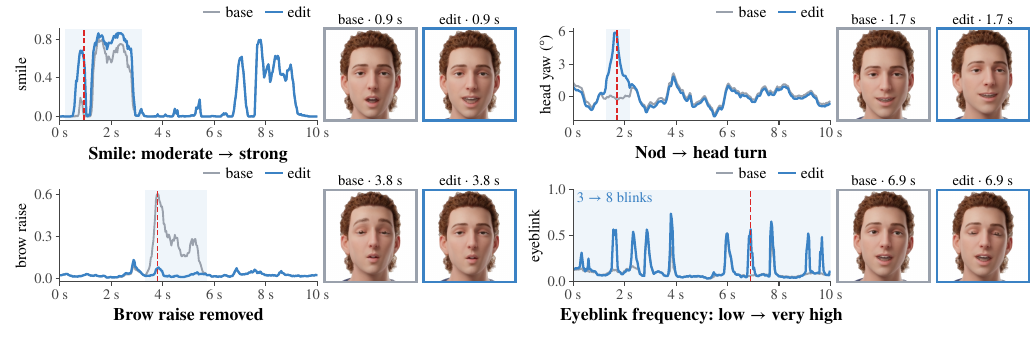}
\caption{\textbf{Changing one instruction.} Same character, notes (\cref{fig:teaser}) and sample; each panel changes one instruction, covering all three layers. Curves: generator output (gray: base; blue: edit); the shaded span is the changed instruction, and the frame at the largest difference is rendered on the right.}
\label{fig:edit}
\end{figure*}

\subsection{Director Language}
\label{sec:language}

The director language describes motions that people can name rather than individual channels. We divide the channels into 9 motion families: eyelids, squint, brow raise, frown, smile, mouth tension, head rotation, head translation and gaze; the channels of family $f$ are denoted by $\mathcal C(f)$. Each set of director notes consists of a clip header and three layers (\cref{fig:teaser}):
\begin{equation}
\begin{gathered}
N=(h,\;\mathcal T,\;\mathcal S,\;\mathcal E),\qquad \mathcal T=\{(\ell_k,\,v_k)\},\\
s=(f_s,\tau_s,\ell_s,v_s)\in\mathcal S,\qquad e=(f_e,\tau_e,\ell_e,A_e)\in\mathcal E
\end{gathered}
\label{eq:notes}
\end{equation}
The clip header $h$ records the neutral value (rest) of each channel and the reference orientation of the head; states and events are expressed as deviations from this baseline. Tendencies $\mathcal T$ describe the style of the whole clip through seven statistics measured over it (eyeblink frequency, head energy, head motion frequency, gaze wandering, gaze shifting frequency, upper face energy and mouth amplitude), each divided into five levels (\texttt{very\_low} \dots{} \texttt{very\_high}) by its quintiles on the training set; in a long clip, a sentence that departs from these defaults can carry its own level (0.9\% of validation clips). States $\mathcal S$ describe expressions lasting at least 1\,s, and events $\mathcal E$ are rapid rises and falls overlaid on the state, such as a blink, a nod or a quick glance. Each state or event has a family $f$, an intensity level $\ell$ (slight, moderate or strong) and four time points $\tau=(\tau_0\le\tau_1\le\tau_2\le\tau_3)$: onset, settled, release and end for a state, and onset, apex start, apex end and offset for an event. The value $v$ of a state is its value on the plateau; the amplitude $A$ of an event is the extra deviation above the state at the apex, and an event also carries a kind and a direction.

Only events are hard constraints: on the frames and channels they cover, the generator must reproduce them within a per-channel tolerance; states and tendencies are soft conditions that the generator follows as closely as possible (\cref{sec:codec,sec:generator}). \Cref{fig:edit} shows the output when one instruction is changed.

The same vocabulary is used at three densities. Full notes are the complete output of the encoder; they are randomly masked into training conditions and serve as the reference for evaluation. Standard and lite notes are sparser versions of the full notes, made by fixed rules.

\subsection{Encoder and Decoder}
\label{sec:codec}

Both are rule-based and training-free, with constants calibrated once on the training set.

\begin{figure*}[t]
\centering
\includegraphics[width=\textwidth]{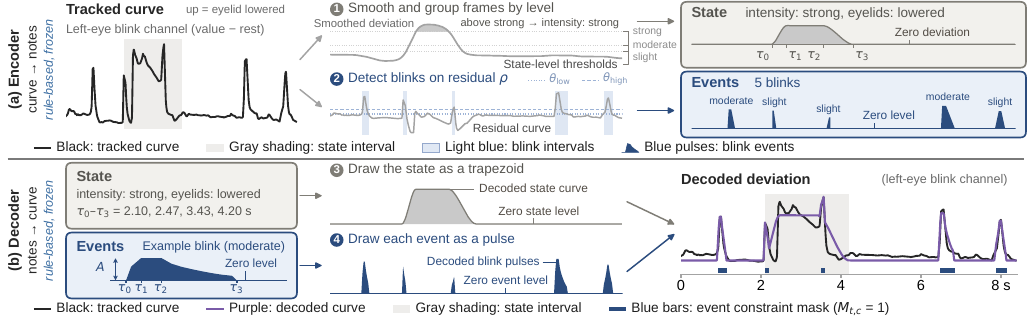}
\caption{\textbf{Encoder and decoder} (eyelid family, 8.4\,s of a training clip; gray: states, blue: events). \emph{(a)} Smoothed frames are grouped by level into a state (1); the remainder is thresholded into events (2). \emph{(b)} States are drawn as trapezoids (3) and events as pulses (4); their sum is $\hat x$ (purple) over the tracked curve (black), with the constrained entries $M$ (dark-blue bars).}
\label{fig:encdec}
\end{figure*}

\paragraph{Decoder.} Each state is reconstructed as a trapezoid $\phi_{\tau}(t)$ with a smooth onset and offset, and
each event as a pedestal-plus-exponential-peak template $y_e(t)$ (\cref{fig:encdec}b).
The full director notes are decoded as
\begin{equation}
\hat x_t=\text{rest}+\sum_{s\in\mathcal S}v_s\,\phi_{\tau_s}(t)\,\mathbf u_{f_s}
+\sum_{e\in\mathcal E}A_e\,y_e(t)\,\mathbf u_{f_e}
\label{eq:decode}
\end{equation}
where $\mathbf u_f$ contains the weights of the channels in family $f$. When only levels are specified, $v_s$ and $A_e$
take the representative training-set value of that level.

The cells covered by events form the constraint mask $M\in\{0,1\}^{T\times 63}$:
\begin{equation}
\begin{gathered}
M_{t,c}=\mathbf{1}\big[\exists\,e\in\mathcal E:\ \tau^e_0\le t<\tau^e_3,\ c\in\mathcal C(f_e)\big],\\
X_\text{obs}=\hat X\odot M
\end{gathered}
\label{eq:mask}
\end{equation}
where $c$ indexes the 63 channels of $x_t$. The constrained values $X_\text{obs}$ are the targets on the constrained entries,
taken from the decoded output $\hat X$ of the given notes.

\paragraph{Encoder.} The encoder takes only three inputs: the tracked 63-dimensional curves, the script timing, and a set of constants calibrated once on the training set. It extracts the notes in order of dependency, starting with the clip header and the states: it heavily smooths the deviation from the neutral value into $\tilde d$, computes the state value of each timeline with a fixed combination rule $g_f$, and assigns an intensity level $\ell_t$ to each frame:
\begin{equation}
\begin{gathered}
\tilde d=\mathcal G_{\sigma=5}\ast\operatorname{med}_{31}(x-\text{rest}),\qquad s^{(f)}_t=g_f(\tilde d_t),\\
\ell_t=\operatorname{sign}\big(s^{(f)}_t\big)\cdot\#\big\{k:\ e^f_k\le|s^{(f)}_t|\big\}
\end{gathered}
\label{eq:level}
\end{equation}
where $\#$ counts the boundaries that $|s^{(f)}_t|$ exceeds, and $|\ell_t|\in\{0,1,2,3\}$ corresponds to neutral, slight, moderate and strong. It then finds the events: it decodes the extracted states into $\hat s$ (\cref{eq:decode}) and detects events with hysteresis on the residual $\rho=x-\text{rest}-\hat s$; for a channel of family $f$, a segment counts as an event if $\rho$ stays above $\theta^f_\text{low}$, peaks above $\theta^f_\text{high}$ and lasts between $d^f_\text{min}$ and $d^f_\text{max}$ frames, and its amplitude $A_e$ is the residual at the first peak frame. Finally, it computes the tendencies from the states and events.

\paragraph{Properties.} Encoding and decoding are both deterministic. All data-estimated constants are fitted only
once, on the 480 TalkVid~\cite{chen2025talkvid} training clips, and then used unchanged on all three datasets;
RAVDESS~\cite{livingstone2018ravdess} and CREMA-D~\cite{cao2014cremad} never take part in the fitting.

\subsection{Sparse-Notes Generator}
\label{sec:generator}

Given sparse notes and the script, the generator
learns to perform the whole clip: specified events must be achieved, states and tendencies should be reflected, and
unspecified parts should look natural.

\paragraph{Input.} The generator is a diffusion model~\cite{ho2020ddpm}. Its input at each frame concatenates the noised motion, the conditions decoded from the director notes (the constrained values $X_\text{obs}$, the constraint mask $M$, the neutral values, the curve of each state timeline and the tendency values) and script features (word timing features and viseme embeddings learned with the generator), 319 values per frame in total (63 each for the noised motion, $X_\text{obs}$, $M$ and the neutral values; 40 for states and tendencies; 27 for the script). Word timing comes from word-level timestamps of faster-whisper large-v3 on the training videos, and visemes from grapheme-to-phoneme conversion (g2p\_en, pypinyin) with a fixed viseme table; audio is used only for this step, never by the generator. We write the conditions given by the director notes and the script $w$ together as $\mathbf c(N,w)$; neither audio nor ground-truth motion is used.

\paragraph{Network.} A 6-layer Transformer~\cite{vaswani2017attention} with width 256, 4 attention
heads and 5.36M parameters.

\paragraph{Training.} The key to training is that the generator sees notes at every degree of sparsity, as in masked modeling~\cite{he2022mae} and training with random partial keyframes~\cite{cohan2024condmdi} or descriptions~\cite{tessler2024maskedmimic}. Each sample is randomly masked by a fixed rule $\tilde N\sim\mathcal R(N)$: we choose a density (full, standard or lite notes, or tendencies only) and how many entries to keep at that density, drop all values with probability 0.5 to keep only the levels, and independently drop each of the three layers entirely with probability 0.15. The ground-truth motion is clipped to within $\pm\delta$ of the decoded values on the constrained entries and then
normalized per channel, yielding $d_0$. At step $i$, noise is added to obtain $z_i$, and the network predicts $v$ using
v-parameterization~\cite{salimans2022progressive}:
\begin{equation}
\begin{gathered}
z_i=\alpha_i d_0+\sigma_i\epsilon,\qquad v=\alpha_i\epsilon-\sigma_i d_0,\\
\mathcal L=\mathbb E_{\tilde N\sim\mathcal R(N),\,i,\,\epsilon}\,\big\|v_\theta(z_i,i,\mathbf c(\tilde N,w))-v\big\|^2
\end{gathered}
\label{eq:loss}
\end{equation}
where $\epsilon\sim\mathcal N(0,I)$ is the noise, and $\alpha_i$ and $\sigma_i$ follow the cosine noise schedule;
$v_\theta$ is the network's prediction of $v$. The training target, however, is still
\emph{the ground-truth motion of the whole clip}, so following the notes alone is not enough: what is specified
must be achieved, and what is left unspecified must be filled in with the motion a real person would make given the
script, tendencies and context, rather than left at the neutral values. Since the constrained entries are also included in the
loss, the model learns to execute the specified events itself, while the projection below guarantees that the constrained entries stay within tolerance; whether events are recovered is measured separately (\cref{sec:exp_main}). Without projection, event recall drops from 0.744 to 0.676 at similar FMD.

\paragraph{Blink masking.} Blinks remaining in the condition are dropped with probability 0.9 but kept in the
training target. The model thus learns that ``no blink specified'' does not mean ``no blinking''; how often to blink
is instead controlled by the eyeblink frequency tendency.

\paragraph{Sampling and hard constraints by projection.} Sampling uses 25-step DDIM~\cite{song2021ddim} ($\eta=0$) without
classifier-free guidance. At each step, we compute the clean motion predicted by the model,
$\hat d_0=\alpha_i z_i-\sigma_i\hat v$, project it on the constrained entries to within $\pm\delta$ of the decoded values, similarly to the known-region replacement of RePaint~\cite{lugmayr2022repaint}, and
use the projected $\hat d_0$ in the next step:
\begin{equation}
\begin{gathered}
\Pi_\delta(y)_{t,c}=\begin{cases}\operatorname{clip}_{[\hat x_{t,c}-\delta_c,\,\hat x_{t,c}+\delta_c]}(y_{t,c})
& M_{t,c}{=}1\\ y_{t,c} & M_{t,c}{=}0\end{cases}\\
\Rightarrow\ \big|x_{t,c}-\hat x_{t,c}\big|\le\delta_c\quad\forall\,(t,c):\ M_{t,c}=1
\end{gathered}
\label{eq:projection}
\end{equation}
where $\delta_c$ is the tolerance of channel $c$, fixed per family in the vocabulary (0.02--0.05 on the $[0,1]$ blendshape scale and $1^\circ$ for head rotation), and $x$ is the generated
output. The generated values on every constrained entry stay within $\delta_c$ of the decoded event, regardless of
what the model learns.

\section{Experiments}
\label{sec:experiments}

\begin{figure*}[t]
\centering
\includegraphics[width=0.85\textwidth]{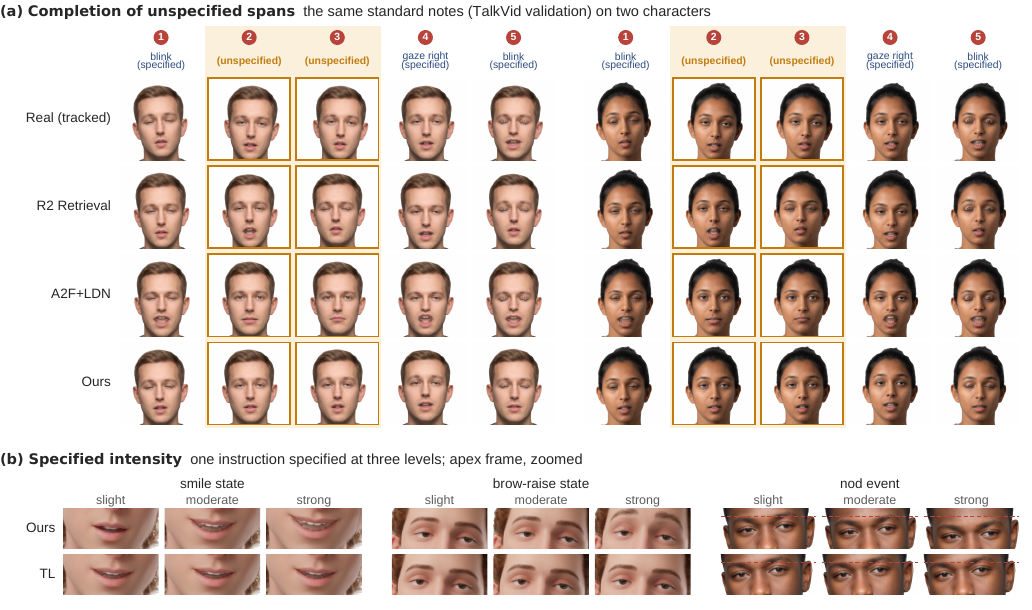}
\caption{\textbf{Qualitative results} (videos online). \emph{(a)} Completion of unspecified spans: the same standard notes of a TalkVid validation clip on two characters; above each column, what the notes specify at that moment. Retrieval pastes mismatched blinks (\eg, moment 3) and A2F+LDN keeps the head and gaze still. \emph{(b)} One instruction specified at three levels (apex frame, zoomed; nod: dashed line at the eye height before the nod): ours follows the level; TL gets one input for all levels.}
\label{fig:qual}
\end{figure*}

\subsection{Setup}
\label{sec:exp_setup}\label{sec:eval}

\paragraph{Data.} Generators are trained on the training split of TalkVid, RAVDESS and CREMA-D (6684 clips) and evaluated on the validation split (1485 clips: CREMA-D 1142, RAVDESS 240, TalkVid 103); the splits are speaker-disjoint. The full notes of each clip are reduced to full, standard and lite notes, and all methods receive the same notes. Results are averaged within each dataset and across datasets, on standard notes unless stated otherwise.

\paragraph{Cycle-consistency evaluation.} The deterministic encoder of \cref{sec:codec} re-encodes the generated motion, and the result is compared with the input notes layer by layer, so no learned evaluator is needed. The encoder is frozen before the generator is trained and is never used during sampling; on real motion it recovers 0.996 of the specified events (\cref{tab:main}), so its readings are not specific to generated output. Its agreement with human annotation is reported in \cref{sec:exp_codec}, and human judgments of the output in \cref{sec:exp_user}. Event recall is the fraction of specified events that are detected, and event intensity agreement is the fraction of matched events recovered at the specified level. State execution is the fraction of frames with a specified non-neutral level whose level and direction are both recovered; tendency execution is the fraction of tendency levels recovered, scored per statistic and sentence segment so that sentence-level levels count as well. Against single-layer timelines we also report the Timeline Alignment Score (TAS) of Timeline Control~\cite{ma2025timeline}, which scores only whether and when an action occurs. Naturalness is measured by a velocity Fr\'echet Motion Distance (FMD): the Fr\'echet distance of~\cite{heusel2017fid}, computed on unconstrained entries over PCA-projected 30-frame windows.

\paragraph{Our model and baselines.} Our generator (\cref{sec:generator}) is trained with 3 seeds, whose results are averaged; the projection is applied at every DDIM step and once more after the head smoothing that follows sampling. Three training-free baselines copy the decoded notes on the constrained entries and differ elsewhere: R0 keeps the neutral value, R1 interpolates, and R2 pastes the training clip whose start and end match best. A2F represents audio-driven methods: it synthesizes speech with Kokoro~\cite{hexgrad2025kokoro} and drives Audio2Face-3D~\cite{nvidia2025audio2face}, with head and gaze neutral; A2F+LDN overwrites the constrained entries with the decoded notes.

\paragraph{Other conditioning formats.} Generators with the same architecture, data and training (3 seeds each) take a single-layer format instead: TL, our 45 binary timelines over the LDN vocabulary, without intensity or tendencies and with states and events on one flat timeline; or TC16, the 16 labels of Timeline Control~\cite{ma2025timeline}. TC16 is a controlled format comparison rather than a reproduction of Timeline Control: its labels are derived from our encoder, and its generator shares our architecture and data. Both are converted deterministically from the same notes, except in the emotion experiment (\cref{sec:exp_timeline}) and in \cref{sec:exp_llm}, where the language model authors every format. Language models also author keyframes: blendshapes (BS, linearly interpolated) or FACS action units (AU, 24 units plus head and gaze, mapped to ARKit blendshapes).

\subsection{Encoder--Decoder Fidelity}
\label{sec:exp_codec}

The encoder provides both the training labels and the evaluation readings, so it must recover what the decoder produces. Decoding the full notes and re-encoding the curves gives an event F1 of 0.88 and a per-frame state agreement of 0.89; for standard notes, decoding alone gives an event recall of 0.84, compared with 0.996 for human motion.

Refitting the encoder constants on RAVDESS or CREMA-D instead of TalkVid changes the encoded notes only slightly: event F1 and state agreement are 0.95--0.97, within 0.004--0.03 of a noise floor from refitting on two speaker-disjoint halves of one dataset.

\paragraph{Agreement with human annotation.} We marked the approximate position of every short event in 24 validation clips on the raw video, blind to the encoder. Matching each annotated event to an encoded event of the same family whose centre lies within 0.5\,s, the encoder reaches a micro-F1 of 0.53 (95\% CI 0.47--0.59) and a macro-F1 of 0.44 (0.36--0.52). Agreement is high for blinks (0.88) and moderate for head and gaze motion (0.58 and 0.53); for brow raise and squint the encoder reports many unmarked events.

\paragraph{Comparison with a vision--language reader.}\label{sec:exp_vlm} On 24 silent validation renders, tracking + encoder recovers 59\% of specified events, Gemini 3.1 Pro 18\%.

\subsection{Naturalness of Completion and Execution}
\label{sec:exp_main}\label{sec:exp_exec}

\Cref{tab:main} compares completion and execution with the baselines of \cref{sec:exp_setup}. See \cref{fig:qual} for examples.

\begin{table}[t]
\centering
\footnotesize
\setlength{\tabcolsep}{1.5pt}
\begin{tabular}{@{}lccccc@{}}
\toprule
Method & FMD $\downarrow$ & Seam $\downarrow$ & Event $\uparrow$ & State $\uparrow$ & Tendency $\uparrow$ \\
\midrule
R0 Neutral & 8.00 & 0.40 & \best{0.868} & 0.042 & \second{0.514} \\
R1 Interpolation & 8.23 & \second{0.29} & 0.781 & 0.047 & \second{0.514} \\
R2 Retrieval & \second{1.62} & 0.33 & 0.569 & \second{0.149} & 0.483 \\
A2F & 9.12 & 1.89 & 0.063 & 0.068 & 0.490 \\
A2F+LDN & 6.64 & 0.72 & \second{0.816} & 0.109 & 0.509 \\
Ours & \best{1.00} & \best{0.09} & 0.744 & \best{0.326} & \best{0.615} \\
\textit{Human (reference)} & --- & --- & \textit{0.996} & \textit{0.778} & \textit{0.991} \\
\bottomrule
\end{tabular}
\caption{Main comparison on standard notes (macro average over 3 datasets). FMD on unconstrained entries, normalized so that Ours is 1. Seam: acceleration where constrained and unconstrained entries meet, relative to real motion (0 = as smooth as real). Human: real motion re-encoded, the ceiling (below 1 for states because standard notes merge states). Results are colour-coded by \best{best} and \second{second best} per column.}
\label{tab:main}
\end{table}

Our model has the lowest FMD and seam and executes states and tendencies best (\cref{tab:main}). Its event recall is lower than that of R0, R1 and A2F+LDN, but recall is limited by the encoder and decoder (0.84 from decoding alone, \cref{sec:exp_codec}; a reference level, not a strict bound); R0 exceeds it only by placing events on a still background, at $8\times$ our FMD. Unspecified spans receive events at a human-like rate (141 vs.\ 166 per minute).

\subsection{Comparison with Single-Layer Timelines}
\label{sec:exp_timeline}\label{sec:exp_control}\label{sec:exp_intent}

\Cref{sec:exp_main} does not isolate the benefit of intensity and layering. The same standard notes of 299 validation clips with a script are converted into TC16, TL and LDN, each fed to the generator trained on that format, and the outputs are re-encoded (\cref{tab:timeline}).

\begin{table}[!htb]
\centering
\footnotesize
\setlength{\tabcolsep}{4pt}
\begin{tabular*}{\linewidth}{@{\extracolsep{\fill}}lcccc@{}}
\toprule
 & \textit{No notes} & TC16 & TL & Ours \\
\midrule
TAS16 macro $\uparrow$ & 0.27 & 0.441 & \second{0.468} & \best{0.471} \\
TAS16 micro $\uparrow$ & 0.44 & \second{0.566} & 0.564 & \best{0.590} \\
TAS45 macro $\uparrow$ & 0.16 & 0.245 & \second{0.321} & \best{0.361} \\
Event intensity agreement $\uparrow$ & 0.40 & 0.492 & \second{0.536} & \best{0.823} \\
Amplitude error $\downarrow$ & 0.45 & \second{0.37} & 0.38 & \best{0.12} \\
State level agreement $\uparrow$ & 0.18 & 0.228 & \second{0.241} & \best{0.277} \\
\bottomrule
\end{tabular*}
\caption{The same notes in three formats, each with its own generator (299 validation clips $\times$ 3 seeds). TL and TC16 enter as binary label tracks without constrained entries or projection; \textit{No notes}: no conditions. TAS16, TAS45 and the other metrics are defined in \cref{sec:exp_setup}. Results are colour-coded by \best{best} and \second{second best} per column.}
\label{tab:timeline}
\end{table}

LDN matches or exceeds the timelines on every TAS score, so carrying more information does not cost timing. Timeline labels carry no intensity, so their generators fall back on typical human behavior: event intensity agreement stays close to that with no notes, with about three times the amplitude error of LDN. Without projection, our generator alone still reaches 0.67 and 21\% (TL: 0.54, 38\%); TL and TC16 have relative FMD 1.06 and 1.08.

\begin{figure}[!htb]
\centering
\includegraphics[width=\linewidth]{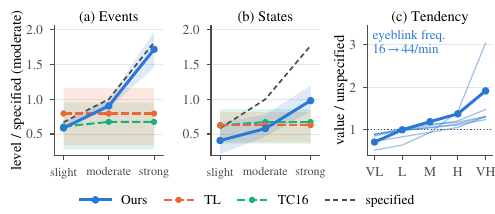}
\caption{Controllability (299 validation clips $\times$ 3 seeds). Good performance follows the dashed line. \textbf{(a)} Events and \textbf{(b)} states: movement relative to the moderate level; dashed: specified target; bands: range. Flat: level ignored (TL gets one input for all levels; TC16 separates two smile levels). \textbf{(c)} Recovered tendency relative to nothing specified (1 = unchanged); should rise from VL to VH. Thick: eyeblink frequency.}
\label{fig:control}
\end{figure}

\paragraph{Intensity.} We first specify a single state or event (6 states, 4 events) at the slight, moderate or strong level and leave the rest empty (\cref{fig:control}). With LDN, the response to all 10 instructions rises monotonically with the level: events reach 83--100\% of the specified value, and states 42--94\%. Levels also control the style of the whole clip: we specify only one tendency, at a level from very low to very high (\eg, a very high eyeblink frequency), and nothing else. All seven tendencies follow the specified level monotonically; one word lifts blinks from 16 to 44 per minute.

\paragraph{Emotion.} Intensity also helps with emotion. For 184 RAVDESS and CREMA-D validation clips, Opus (\cref{sec:exp_llm}) authors notes requesting a target emotion, with intensity ``a little'' or ``very'' for 148 of them; a classifier trained on the training-set actors judges the output. LDN reaches an accuracy of 0.724 on RAVDESS and 0.592 on CREMA-D, against 0.604 / 0.556 for TL and 0.568 / 0.525 for TC16 (real validation clips: 0.703 / 0.642). From ``a little'' to ``very'', the probability of the target emotion rises by 0.13--0.15 with LDN, but by only 0.03--0.05 with timelines.

When all three layers are specified together, each layer is executed no worse than when specified alone; \eg, event recall rises from 0.686 to 0.746.

\subsection{User Study}
\label{sec:exp_user}

Participants choose between two silent clips on the same LAM avatar: the one that moves more naturally (ours against R2, A2F+LDN or real motion) or the one with the stronger action (one instruction specified as slight or strong, or a very low or very high eyeblink frequency). We keep 30 of 40 submissions, excluding 6 with a median below 3\,s per question and 4 repeats; the stimuli were selected by chance from a larger rendered pool, and the protocol is in the supp.

\begin{table}[!htb]
\centering
\footnotesize
\setlength{\tabcolsep}{2.5pt}
\begin{tabular*}{\linewidth}{@{\extracolsep{\fill}}c@{\hspace{3pt}}lcccl@{}}
\toprule
 & & Rate & 95\% CI & $p$ & Result \\
\midrule
\multirow{3}{*}{\rotatebox[origin=c]{90}{\tiny\textit{Natural.}}} & Ours vs.\ A2F+LDN & 0.86 & [0.78, 0.92] & $<$0.001 & \textbf{better} \\
 & Ours vs.\ R2 & 0.67 & [0.59, 0.74] & $<$0.001 & \textbf{better} \\
 & Ours vs.\ real & 0.45 & [0.36, 0.55] & 0.33 & n.s. \\
\midrule
\multirow{3}{*}{\rotatebox[origin=c]{90}{\tiny\textit{Intensity}}} & Ours (smile, brow, nod) & 0.84 & [0.76, 0.92] & $<$0.001 & \textbf{perceived} \\
 & TL (same instructions) & 0.51 & [0.41, 0.60] & 0.92 & chance \\
 & Ours, eyeblink freq. & 0.87 & [0.73, 0.97] & $<$0.001 & \textbf{perceived} \\
\bottomrule
\end{tabular*}
\caption{User study (30 participants, two-alternative forced choice; 95\% CI by participant bootstrap; exact binomial $p$ against 0.5; about 130 trials per naturalness comparison). Ours is significantly preferred over both baselines and not significantly different from real motion, and its specified intensity is perceived; TL gets one input for slight and strong, so 0.5 is expected.}
\label{tab:user}
\end{table}

Ours is preferred over A2F+LDN and retrieval ($p<0.001$, exact binomial test) and not distinguished from real motion ($p=0.33$; \cref{tab:user}); the strong version is identified with ours (0.84) but not with TL (0.51), and so is the very high eyeblink frequency (0.87), a judgment made by people rather than by our encoder.

\subsection{\texorpdfstring{Authoring LDN with Language Models}{Authoring LDN with Language Models}}
\label{sec:exp_llm}

\begin{table}[!htb]
\centering
\footnotesize
\setlength{\tabcolsep}{2pt}
\begin{tabular*}{\linewidth}{@{\extracolsep{\fill}}lcccc@{}}
\toprule
 & \multicolumn{2}{c}{\textit{Opus}} & \multicolumn{2}{c}{\textit{Haiku}} \\
\cmidrule(lr){2-3}\cmidrule(l){4-5}
Format & Validity $\uparrow$ & FMD $\downarrow$ & Validity $\uparrow$ & FMD $\downarrow$ \\
\midrule
Ours & 0.990 / 0.963 & \second{1} & 0.936 / 0.656 & 1 \\
TL & \best{1.000} / \second{0.970} & \best{0.95} & 0.967 / \second{0.926} & \best{0.91} \\
TC16 & \best{1.000} / \best{0.993} & 1.01 & 0.987 / \best{0.963} & \second{0.97} \\
BS keyframes & \second{0.997} / 0.692 & 3.03 & \second{0.997} / 0.398 & 3.44 \\
AU keyframes & \best{1.000} / 0.642 & 3.08 & \best{1.000} / 0.425 & 3.52 \\
\bottomrule
\end{tabular*}
\caption{Formats authored by language models (299 clips). Validity: after repair / strict; FMD: all-frame, vs.\ Ours. Results are colour-coded by \best{best} and \second{second best} per column.}
\label{tab:llm}
\end{table}

Given only the clip length and the timed script, Claude Opus 5.5 and Claude Haiku 5.5 author all five formats without fine-tuning, using one shared prompt with three examples; Gemini 3.1 Pro is in the supp. The models receive the real word timing, so that only the authoring of the notes is evaluated; in use, any TTS can supply it.

As shown in \cref{tab:llm}, both models author LDN validly after repair; keyframes become valid only after repair and are about three times less natural, and timelines are as valid and natural as LDN. The advantage of LDN lies in intensity and emotion (\cref{sec:exp_timeline}), not in ease of authoring or naturalness.

\subsection{Ablation Study}
\label{sec:exp_ablation}

\begin{table}[!htb]
\centering
\footnotesize
\setlength{\tabcolsep}{3pt}
\begin{tabular*}{\linewidth}{@{\extracolsep{\fill}}c@{\hspace{3pt}}lcccc@{}}
\toprule
 & Configuration & FMD $\downarrow$ & Recall $\uparrow$ & Mouth $\uparrow$ & Blink $\downarrow$ \\
\midrule
\multirow{3}{*}{\rotatebox[origin=c]{90}{\tiny\textit{Cond.}}} & Script only & 1.17 & 0.172 & \second{0.232} & \best{0.204} \\
 & LDN only & \best{0.97} & \second{0.827} & 0.115 & \second{0.230} \\
 & Ours (LDN + script) & \second{1.00} & \best{0.830} & \best{0.299} & 0.263 \\
\midrule
\multirow{2}{*}{\rotatebox[origin=c]{90}{\tiny\textit{Mask}}} & No blink masking & 1.08 & 0.745 & 0.222 & 0.248 \\
 & Ours ($p=0.9$) & \best{1.00} & 0.745 & 0.222 & \best{0.175} \\
\bottomrule
\end{tabular*}
\caption{Ablation (validation set, 3 seeds; groups differ in setting). Mouth: jaw corr.\ with real; Blink: blink error. Results are colour-coded by \best{best} and \second{second best} per column within each group.}
\label{tab:ablation}
\end{table}

In \cref{tab:ablation}, the script alone recalls 0.17 of specified events and LDN alone barely follows the real jaw (0.12); together they reach 0.83 and 0.30. Blink masking lowers the blink error from 0.248 to 0.175 and FMD from 1.08 to 1.00.

\section{Conclusion and Limitations}
\label{sec:conclusion}\label{sec:limitations}
We presented LDN, a layered, sparse language for directing facial performance, and a generator that executes specified notes and completes unspecified spans. Sustained states reach 42--94\% of their specified values, while events reach 83--100\%. The encoder that labels the training data also scores the output, and it marks more brow and squint events than human annotators. Without audio, the mouth follows word timing; when audio exists, an audio-driven lip model can drive the speech channels (videos online).

\clearpage
\section*{LLM Usage Statement}

We used Claude Code for language polishing and proofreading, to help draft and shorten parts of the text, and to assist with implementing the data, training, evaluation and figure scripts and with running the experiments. All AI-assisted material was reviewed and validated by the authors. The method, the experimental design and the conclusions are the work of the authors, who take full responsibility for this manuscript. Separately, language models are subjects of our experiments rather than components of the LDN system: Claude Opus 5.5 and Claude Haiku 5.5 author the director notes evaluated in \cref{sec:exp_llm}, and Gemini 3.1 Pro serves as the vision--language reader in \cref{sec:exp_vlm} and as an additional author in the supp. The generator, encoder and decoder of the LDN system contain no language model. The portrait photographs in the figures were generated from text descriptions with Gemini 3 Pro Image and depict no real person.

{
    \small
    \bibliographystyle{ieeenat_fullname}
    \bibliography{main}
}

\clearpage
\setcounter{dbltopnumber}{2}
\setcounter{topnumber}{3}
\setcounter{bottomnumber}{2}
\setcounter{totalnumber}{6}
\renewcommand{\topfraction}{0.9}
\renewcommand{\bottomfraction}{0.6}
\renewcommand{\dbltopfraction}{0.9}
\renewcommand{\textfraction}{0.08}
\renewcommand{\floatpagefraction}{0.8}
\renewcommand{\dblfloatpagefraction}{0.8}
\appendix
\maketitlesupplementary
\renewcommand{\theHsection}{supp.\Alph{section}}
\renewcommand{\theHtable}{S\arabic{table}}
\renewcommand{\theHfigure}{S\arabic{figure}}
\renewcommand{\theHequation}{S\arabic{equation}}
\renewcommand{\thesection}{\Alph{section}}
\renewcommand{\thetable}{S\arabic{table}}
\renewcommand{\thefigure}{S\arabic{figure}}
\renewcommand{\theequation}{S\arabic{equation}}
\setcounter{section}{0}\setcounter{table}{0}\setcounter{figure}{0}\setcounter{equation}{0}

\begin{strip}
\centering
\includegraphics[width=\textwidth]{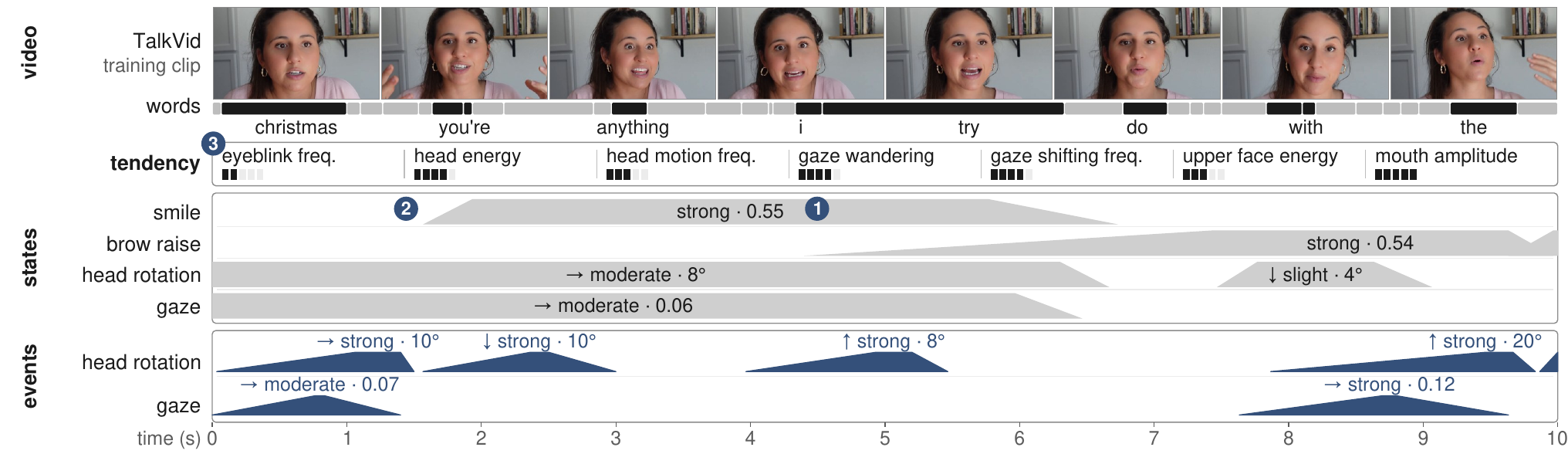}
\captionsetup{hypcap=false}
\captionof{figure}{\textbf{Director notes for 10\,s of a training clip} (standard notes, extracted by the encoder). \emph{Top:}
video frames (subtitles show the word being spoken) and the timing of every word. Tendencies describe the whole clip in
five levels (the number of filled squares is the level). Values are in blendshape units or degrees; for head-rotation
events, the axis with the largest absolute value is shown. \textcircled{\scriptsize 1}~Intensity: the smile is specified
as ``strong\,$\cdot$\,0.55''; \textcircled{\scriptsize 2}~shape: its onset takes 0.37\,s;
\textcircled{\scriptsize 3}~layering: the style of the whole clip is specified once, in the tendency layer. For space,
states are shown for only 4 families; the notes of this window also contain states of squint, mouth tension and head translation. All tendencies and events in the window are shown.}
\captionsetup{hypcap=true}
\label{fig:supp_notes}
\end{strip}
\vspace{2pt}

\noindent This document gives the formulas, implementation details and further results referred to in the main
paper. Section, table, figure and equation numbers without the prefix ``S'' or a letter refer to the main paper.

\section{Director Language Details}
\label{sec:supp_language}

\paragraph{Families and channels.} \Cref{fig:supp_notes} shows complete standard notes of one clip, and \cref{tab:supp_families} lists the channels of each family. Every channel
belongs to at most one family. %
The opening and closing of the mouth during speech does not belong to any family, but is determined by the script. %
These 14 speech channels are driven by the script features of the generator (\cref{sec:supp_generator}).
Channels that cannot be measured, such as \texttt{mouthFrown} and \texttt{cheekPuff}, currently do not belong to any family; they can be added as new families with a better tracker. %
At present, we only include motions that can be stably measured by the tracker. The family table can be extended. %

\paragraph{Full grammar.} Eq.~(1) is simplified. The full tendency and event entries are
\begin{equation}
\mathcal T=\{(k,\pi,\ell_k,v_k)\},\qquad e=(f_e,\tau_e,\ell_e,A_e,\kappa_e),
\end{equation}
where $k$ is the statistic, $\pi$ the time span (the whole clip or one sentence) and $\kappa_e$ the kind and
direction of the event.
Each tendency entry specifies the level $\ell_k$ and an optional value $v_k$ of the $k$-th statistic over a time span $\pi$. %
Each statistic has a default level over the whole clip; when a sentence needs a different style, its level can be overridden for that sentence alone. %
The kind and direction $\kappa$ can also be left unspecified, and so can the side; if the side is unspecified, the motion is symmetric. %
For vector families, the amplitude has per-axis components. A state can also carry a value, a direction and a side,
and its onset and offset speeds are given by its four time points.

\begin{table*}[t]
\centering
\footnotesize
\setlength{\tabcolsep}{6pt}
\renewcommand{\arraystretch}{1.25}
\begin{tabular}{@{}l >{\raggedright\arraybackslash}p{0.46\textwidth} >{\raggedright\arraybackslash}p{0.36\textwidth}@{}}
\toprule
Family & Channels (ARKit~\cite{apple2017arkit} names) & Directions and kinds \\
\midrule
Eyelids & eyeBlink L/R minus eyeWide L/R & $+$ lowered, $-$ widened; event kind: blink \\
Squint & eyeSquint L/R & --- \\
Brow raise & browInnerUp, browOuterUp L/R & --- \\
Frown & browDown L/R & --- \\
Smile & mouthSmile L/R & --- \\
Mouth tension & lip pressor: mouthPress L/R; dimpler: mouthDimple L/R; chin raiser: mouthShrugLower;
upper lip raiser: mouthUpperUp L/R, mouthShrugUpper & four kinds: lip pressor, dimpler, chin raiser, upper lip raiser \\
Head rotation & rotation vector, 3 axes & nod (down / up), turn (image left / right), tilt (image cw / ccw) \\
Head translation & 3 axes & image left / right, up / down, forward / back \\
Gaze & pitch and yaw & --- \\
\midrule
\textit{Speech} & jawOpen, jawLeft/Right, mouthFunnel, mouthPucker, mouthClose, mouthRollLower/Upper, mouthLowerDown L/R,
mouthStretch L/R, mouthLeft/Right & driven by the script \\
\textit{Unused} & cheekPuff, cheekSquint L/R, noseSneer L/R, jawForward, mouthFrown L/R & not measured reliably \\
\bottomrule
\end{tabular}
\caption{Motion families. Speech channels belong to no family and are driven by the script; unused channels are
not measured reliably by the tracker.}
\label{tab:supp_families}
\end{table*}

\paragraph{Rules.} Relaxed faces differ across people: some have naturally raised mouth corners, and some habitually tilt their heads. %
The clip header therefore records each person's neutral values, which are added back during decoding (Eq.~(2)).
In this way, the same ``moderate squint'' means the same amount of change on different people. %
States on the same timeline and events of the same family cannot overlap; however, events of different families can overlap. %
Event durations vary by family, ranging from less than 0.1\,s for a blink to several seconds. %
Repeated motions, such as nodding twice in a row, are expressed as a single note plus a repeat count and a period. %
Direction words are based on the image; for instance, \texttt{image\_left} refers to the left of the image rather than the person's own left. Thus, notes can be authored while watching the video. %
The extreme case of sparsity is empty notes, with nothing specified at all.

\paragraph{Tendencies.} The seven statistics are defined as follows: eyeblink frequency is the number of blinks per
minute; head energy is the average angular speed of head rotation; head motion frequency is the number of nods, head
turns and head tilts per minute; gaze wandering is the proportion of time the gaze deviates from its usual state;
gaze shifting frequency is the number of gaze shifts per minute; upper face energy is the average movement speed of
the eyes and brows; and mouth amplitude is the amplitude of mouth opening and closing during speech. %
Mouth amplitude uses only frames within words, which is why the encoder requires word-level timing. %
A language model does not specify mouth shapes frame by frame; it can only adjust the overall speaking amplitude
through the mouth amplitude tendency.

\paragraph{Densities.} Full notes are the complete output of the encoder; people and language models are not expected to author notes this densely. %
Standard notes, intended for strong off-the-shelf language models, may include values, sides and repeats; lite notes,
intended for small language models, omit them. %
Standard notes contain at most 30 events per minute and at most 6 state segments per family per minute; lite notes
include at most 6 events per minute and at most 2 state segments per family per minute. %
The most salient events are kept (salience = amplitude $\times$ duration). When the states exceed the quota, the two adjacent segments of the same family that are closest in amplitude are merged repeatedly. All tendencies are kept. %

\section{Encoder and Decoder Details}
\label{sec:supp_codec}

\paragraph{Decoder.} Each state is reconstructed as a trapezoid with a smooth onset and offset:
\begin{equation}
\phi_\tau(t)=\begin{cases}\psi(u_\uparrow) & \tau_0\le t<\tau_1\\ 1 & \tau_1\le t<\tau_2\\ \psi(u_\downarrow) & \tau_2\le t<\tau_3\\ 0 & \text{otherwise}\end{cases}
\end{equation}
with $u_\uparrow=\frac{t-\tau_0}{\tau_1-\tau_0}$, $u_\downarrow=\frac{\tau_3-t}{\tau_3-\tau_2}$ and $\psi(u)=3u^2-2u^3$. When two adjacent states share the same transition (the release of the former coincides with the onset of the latter), the curve transitions smoothly from the value of the former to that of the latter and does not return to the neutral value. %
Each event is reconstructed with a pedestal-plus-exponential-peak template:
\begin{equation}
\begin{gathered}
y_e(t)=P(t)+\big(1-P(t)\big)\,e^{-d(t)/\kappa^f},\\
P(t)=b_e+(h_f-b_e)\,\psi(u),
\end{gathered}
\end{equation}
where $d(t)$ is the distance in frames to the apex and $u$ is the normalized time as in the state trapezoid. %
The rising side uses $\kappa^f_\uparrow$ and the falling side uses $\kappa^f_\downarrow$; $y_e=1$ on the apex and $y_e=0$ outside the event. %
The peak rises exponentially to the apex and then decays exponentially from it. %
The pedestal is raised by a smoothstep from $b_e=\min(\theta^f_\text{low}/|A_e|,\,\theta^f_\text{low}/\theta^f_\text{high})$ to a height $h_f$, where $\theta^f_\text{low}$ and $\theta^f_\text{high}$ are the detection thresholds for events (\cref{eq:supp_detect}): during detection, an event counts as started only if the residual exceeds $\theta^f_\text{low}$, so decoding and detection are aligned at the onset. %
The parameters $h_f$, $\kappa^f_\uparrow$ and $\kappa^f_\downarrow$ are calibrated once by family on the training set. %
The time constants do not depend on event duration, so a long event is not stretched into one broad pulse. %
For head rotation, the summation in Eq.~(2) is carried out in rotation-vector space. %
The constrained values $X_\text{obs}$ are the sum of the neutral values, states and events on the constrained
entries, and the other entries are set to zero.
The decoder reads only the fields of the notes and takes no other input, so ground-truth motion cannot leak to the generator through it. %

\paragraph{Encoder: clip header.} The neutral value of one-sided families (\eg, smile and squint) takes the 10th percentile of the clip, \ie, the most relaxed moment; other channels take the median; both use only frames with valid tracking. The reference orientation is the average head orientation over the whole clip, and all head rotations are expressed as rotation vectors relative to it. %

\paragraph{Encoder: states.} A director authors notes by family: ``squint'' is specified as a single motion, although the family may contain several channels. %
Thus, this step needs to compute a state curve $s^{(f)}_t$ for each timeline, which indicates how strongly the motion is performed in each frame; the subsequent level assignment, segmentation and value extraction are all carried out on this curve. %
The combination rule $g_f$ is: one-sided families take the average over their channels; for eyelids, we take the eye-closure value minus the eye-widening value (positive values indicate eye closure, while negative values indicate eye widening); for mouth tension, we take the strongest of the four motions and record its kind; each axis of the head and gaze forms its own timeline, giving 14 timelines in total. %
Taking the average does not remove individual differences: the deviation is computed relative to the person's own neutral value; how the value is distributed among the channels is recovered during decoding by the channel weights $\mathbf u_f$ calibrated on the training set; and the left--right difference is recorded separately as asymmetry (asym / side) and split between the two sides during decoding. %
In Eq.~(4), the window and $\sigma$ are both measured in frames (30\,Hz); the scale of smoothing is about 1\,s.
The median filter removes fast motions such as blinks, which are left to the event layer, and the Gaussian smoothing then rounds off the curve. %
The boundary $e^f_0$ is a hand-set activation threshold (\cref{tab:supp_constants}); $e^f_1$ and $e^f_2$ are the 50th and 85th percentiles of $|s^{(f)}|$ over the frames where $|s^{(f)}|\ge e^f_0$ on the training set; the sign denotes the direction. %
Setting the boundaries by percentiles gives the levels a consistent meaning relative to the people in the training set. For example, the strongest 15\% of active frames are labeled strong. %
Consecutive frames with the same level form a segment, and each non-neutral segment is expressed as a state. %
By definition, states last at least 1\,s; thus, segments shorter than 1\,s are merged into the adjacent segment with the closest level. %
The four time points of each state are the frames at which the state value has covered 10\% and 90\% of the way from the plateau of the previous segment to that of the next (the plateau of a neutral segment is zero). %
The value $v_s$ is the average of the state curve on the plateau.

\paragraph{Encoder: events.} We decode the extracted states into $\hat s$ and detect events with hysteresis on the
residual $\rho=x-\text{rest}-\hat s$. For a channel $c$ in family $f$, a segment $[a,b)$ counts as an event only if
\begin{equation}
\begin{gathered}
\rho_{t,c}\ge\theta^f_\text{low}\ \ \forall t\in[a,b),\qquad \max_{t\in[a,b)}\rho_{t,c}\ge\theta^f_\text{high},\\
b-a\in[d^f_\text{min},d^f_\text{max}].
\end{gathered}
\label{eq:supp_detect}
\end{equation}
For families with a direction, events are detected on $+\rho$ and $-\rho$ separately. Overlapping segments on the channels of the same family are merged into one event, and the duration condition is checked on the merged segment. The apex segment is the contiguous segment that contains the first peak and stays above 90\% of the peak value. %
The amplitude $A_e$ is the residual $\rho$ at the first peak frame. The intensity is divided into levels by the tertiles of the amplitudes on the training set. %

\paragraph{Encoder: tendencies.} We split the clip into segments based on sentence boundaries; segments shorter than 3\,s are merged with an adjacent segment, so a segment may contain several sentences. The seven statistics are computed directly as defined. For instance, eyeblink frequency is the number of blink events within the segment divided by its duration. %
In addition, the seven statistics are computed once over the whole clip, which gives the default levels. %
Levels are the quintiles of each statistic on the training set.

\paragraph{Properties and constants.} The same input always yields the same output, and no manual labeling is required. %
All levels (states, events and tendencies) are defined relative to the people in the training set. Once the constants are frozen, the same value falls into the same level in any dataset, so levels are comparable across datasets. %
A dataset that is more active overall will have a higher level distribution as a whole. %
The constants include the state level boundaries, channel weights, event template parameters, amplitude tertiles,
representative amplitudes and tendency quintiles. The hand-set thresholds and the calibrated state boundaries are listed
in \cref{tab:supp_constants}; all constants are calibrated on the 480 TalkVid~\cite{chen2025talkvid} training clips and released with the code.
All constants are calibrated on all 480 TalkVid training clips, \ie, the whole TalkVid training split, with no
further selection. The state level boundaries $e^f_1$ and $e^f_2$ are the 50th and 85th percentiles of $|s^{(f)}|$
over the training frames with $|s^{(f)}|\ge e^f_0$. The channel weights $\mathbf u_f$ of a state are no-intercept
least-squares fits of each channel's smoothed deviation on the state value of its family (for eyelids, closing and
widening separately); gaze uses an $8\times2$ least-squares map from the eight eyeLook channels to the two gaze
offsets, and the channel weights of events are fitted in the same way on the residual at apex frames. The
representative value of each state level is the median $|v|$ of that level; event amplitudes are divided into levels
by the tertiles of $|A|$, and the representative value of each level is its median. The template parameters $h_f$,
$\kappa^f_\uparrow$ and $\kappa^f_\downarrow$ are chosen on a grid ($h$ from 0 to 0.6 in steps of 0.05; $\kappa$ from
0.5 to 16 frames, log-spaced) to minimize the squared error between the template and the residual over the rise and
fall frames of the events detected on the training set. The tendency quintiles are computed over the training clips.
The thresholds $e^f_0$, $\theta^f$, $d^f$ and $\delta$ are set by hand in the vocabulary
(\cref{tab:supp_constants}) and are not estimated from data.

\begin{table*}[t]
\centering
\footnotesize
\setlength{\tabcolsep}{2.2pt}
\begin{tabular}{@{}lcccc@{}}
\toprule
Timeline & $e_0$ / $e_1$ / $e_2$ & $\theta_\text{low}$ / $\theta_\text{high}$ & $d_\text{min}$--$d_\text{max}$ (s) & $\delta$ \\
\midrule
Eyelids & 0.10 / 0.19 / 0.40 & 0.10 / 0.20 & 0.067--1 & 0.05 \\
Squint & 0.10 / 0.16 / 0.25 & 0.10 / 0.20 & 0.2--10 & 0.05 \\
Brow raise & 0.10 / 0.19 / 0.35 & 0.10 / 0.20 & 0.2--10 & 0.05 \\
Frown & 0.06 / 0.13 / 0.29 & 0.06 / 0.12 & 0.2--10 & 0.03 \\
Smile & 0.10 / 0.31 / 0.56 & 0.10 / 0.20 & 0.3--20 & 0.05 \\
Mouth tension & 0.50 / 0.84 / 1.48 & per kind$^\dagger$ & 0.3--10 & per kind$^\dagger$ \\
Head rot.\ x / y / z ($^\circ$) & 3.0 / 4.6 / 7.8, 4.0 / 6.4 / 10.6, 3.0 / 5.0 / 8.4 & 2 / 4 & 0.2--5 & 1 \\
Head transl.\ x / y / z & 0.9 / 1.5 / 2.7, 0.5 / 0.8 / 1.5, 0.7 / 1.3 / 3.0 & --- & --- & --- \\
Gaze pitch / yaw & 0.010 / 0.017 / 0.031, 0.030 / 0.057 / 0.129 & 0.01 / 0.03 & 0.2--10 & 0.005 \\
\bottomrule
\end{tabular}
\caption{Encoder constants. $e_0$: hand-set activation threshold; $e_1$, $e_2$: calibrated state level boundaries
(50th / 85th percentiles of active frames); $\theta$: event hysteresis thresholds; $d$: event duration range;
$\delta$: projection tolerance (non-driver blendshapes of a family: 0.05). Head translation has no events.
$^\dagger$Mouth tension $\theta_\text{low}$ / $\theta_\text{high}$ / $\delta$: press 0.075 / 0.15 / 0.0375, dimple
0.045 / 0.09 / 0.0225, chin raise and upper lip raise 0.10 / 0.20 / 0.05.}
\label{tab:supp_constants}
\end{table*}

\section{Generator Details}
\label{sec:supp_generator}

\paragraph{Input.} Per frame, the 319 input values are: the noised motion (63), the constrained values $X_\text{obs}$
(63), the constraint mask $M$ (63), the neutral values (63), the curve and left--right asymmetry of each state timeline, with a flag indicating whether the state layer is specified, and the tendency values, with a mask showing which tendencies are specified (40 together), %
and the script features (27).
When the state curves, tendencies and constrained values are zero, the flag and masks distinguish between ``specified as neutral'' and ``left unspecified''. %
The script features are word timing features (\eg, whether the frame is inside a word, word onset and offset, pauses and sentence ends) and viseme embeddings, where the script is converted into 13 viseme classes by g2p~\cite{park2019g2pE}; these only give the mouth a rough rhythm of opening and closing, not exact mouth shapes. %
The script serves two purposes. First, it is a reference for authoring the notes: a director or a language model reads the script to decide what to do on which sentence. Second, it is a condition that sets the timing of the performance for the generator, such as which word to nod on or when to blink at a pause. %

\paragraph{Speech channels.} Speech channels are driven by visemes. Visemes are inferred from the text and assigned
durations by rules, so exact mouth shapes require phoneme-level timing or audio; at deployment, these channels can be
handed over to an existing audio-driven lip model (\cref{sec:supp_mouth}). The speech channels do not belong to any
family; hence, replacing them does not affect the parts controlled by the director notes. Lip sync is not the goal of
this paper, nor do we claim it as a contribution. %

\paragraph{Network and long sequences.} The diffusion timestep embedding is added to every frame at every layer. The training window is 96 frames (3.2\,s). States and tendencies are per-frame input conditions; each window sees the values over its own span, so even a state longer than a window does not need to fall entirely within one window. Long sequences are generated with overlapping windows. At each denoising step, the clean motions predicted by the windows are blended into a single sequence with triangular weights (largest at the center of the window and smallest at its ends) and then projected close to the constrained values (Eq.~(6)). %
Because self-attention operates across frames, unconstrained frames can attend to nearby events, states and the script, and fill in smoothly connected motion. %
The Transformer~\cite{vaswani2017attention} has 6 pre-norm layers (width 256, 4 heads, feed-forward width 1024, dropout 0.1, GELU); the input is
projected by a linear layer with sinusoidal positional encoding, and each layer adds a projection of the
conditions and the timestep embedding to its input.

\paragraph{Training data and masking.} With the encoder, every training video has full director notes. %
There are 6684 clips in the training set, totaling about 5.8 hours. Each of the three datasets contributes one third of the frames, and 8 randomly masked variants are sampled for each clip. %
Tendencies are kept with all statistics or a random subset. When the notes are randomly masked, the masked-out events
and states no longer appear in the condition, and the entries they cover are no longer constrained. %
In this way, the same model can accept both empty notes with no entries and full notes. %
Note that the full notes record the motion that actually happened in this video. %

\paragraph{Script perturbation.} The script timing of each sample is handled in one of three ways: 25\% are retimed at a random speaking rate; in 37.5\%, the boundaries of each word are shifted by a few random frames and a few words are randomly dropped; the remaining 37.5\% keep the real speaker's timing. In this way, the script timing at inference, whether estimated by rules or taken from word-level timestamps of speech synthesis or recognition (TTS/ASR), falls within the range seen in training. %
(Configuration: speaking rate 2.5--4.5 words/s, shift up to 3 frames, word drop probability 0.1.)

\paragraph{Diffusion training.} We add noise at a random level to the ground-truth motion~\cite{ho2020ddpm}, and the network, conditioned on the sparse notes and the script, predicts the velocity target $v$ of the v-parameterization below. At inference, we start from pure noise and denoise step by step. %
Normalization is done by subtracting the mean and dividing by the standard deviation on the training set, yielding $d_0$ with all channels on the same scale. %
The full form of Eq.~(5) (v-parameterization~\cite{salimans2022progressive}) is
\begin{equation}
\mathcal L=\mathbb E_{\tilde N\sim\mathcal R(N),\,i,\,\epsilon}\,\frac{\sum_{t,c}W_{t,c}\big(v_\theta(z_i,i,\mathbf c(\tilde N,w))-v\big)^2_{t,c}}{\sum_{t,c}W_{t,c}},
\end{equation}
where $W_{t,c}\in\{0,1\}$ marks the entries that enter the loss: tracking must be valid, and the channel must not be one excluded by rule. The loss is the average of the squared error over these entries. %
We train with AdamW~\cite{loshchilov2019adamw} (learning rate $2\times10^{-4}$, 500 warm-up steps and cosine decay, weight decay 0.01,
gradient clipping 1.0, EMA 0.9999) for 50k iterations with a batch of 128 windows, with 3 random seeds (about 40 minutes
each on one NVIDIA B200 GPU). The model has 5.36M parameters and is trained from random initialization.

\paragraph{Blink masking.} Although blinks are the most frequent events, people and language models rarely specify them one by one. %
In addition to the random masking above, blinks remaining in the condition are dropped with probability 0.9; the eyeblink frequency in the tendencies is left unchanged. %

\paragraph{Sampling.} The number of DDIM steps and the use of guidance are selected on the validation set from 25, 50 and 100 steps, with or without classifier-free guidance~\cite{ho2022cfg}. %
The tolerances are given in the original scale (\eg, 0.05 for squint and $1^\circ$ for head rotation; \cref{tab:supp_constants}).
Head rotation is clipped per axis in rotation-vector space and then mapped back to a rotation. Finally, a light smoothing is applied to the head ($\sigma_h=0.75$ frames) to remove the small jitter left by per-frame clipping. Smoothing may push values outside the tolerance, so we project again. %
For head rotation, Eq.~(6) holds per axis on the rotation vector, \ie, $|\omega_t-\hat\omega_t|\le\delta_\text{rot}$, where $\omega_t$ is the head rotation vector. %
This guarantee holds in parameter space; whether an event is also \emph{recovered} depends on the surrounding motion (Sec.~4.3). %

\paragraph{Data used for calibration and selection.}
Only the training set is used for all constants of the encoder and decoder (calibrated on the 480 TalkVid training clips),
the intensity tertiles and tendency quintiles, the PCA and velocity statistics of FMD, and the ridge regression of the
renderer and the regularization of the emotion classifier (both chosen by speaker-grouped 5-fold cross-validation
within the training set). The validation set is used to select the generator configuration: conditioning, training
window, training variants, blink masking probability $p$, refinement, DDIM~\cite{song2021ddim} steps and guidance, and head smoothing
$\sigma_h$ (\cref{tab:supp_ablation} in \cref{sec:supp_ablation}). The selection rule was fixed before any result was seen; it compares only
naturalness (FMD on unconstrained entries as the primary metric) on standard notes, macro-averaged over the three
datasets, and never looks at the execution of events, states or tendencies; its two deviations are given in
\cref{sec:supp_ablation}.

Since the main results are also reported on the validation set, we checked whether this selection makes them
optimistic. (i) Execution is not part of the rule and barely changes between candidates: in the blink masking, DDIM
step and head smoothing groups, event recall stays within 0.738--0.749 and state execution within 0.324--0.331, so
the execution results of Tab.~1 and Tab.~2 are not picked on the validation set. (ii) The selected configuration is
often not the one with the lowest validation FMD: 100 DDIM steps give 0.75 against 1.00 for the selected 25 steps, no
head smoothing gives 0.94 against 1.00 for $\sigma_h=0.75$, and LDN alone gives 0.97 against 1.00 for LDN with the
script; in each case a lower validation FMD was given up for seam, W1, head jitter or the mouth. (iii) On the test
speakers, never used for training, calibration or selection (\cref{sec:supp_test}), every comparison keeps its
direction: the state execution margin is $+0.32$ (validation: $+0.32$), tendency execution is 0.716 (0.692), and the
FMD of the baselines relative to ours is larger on the test speakers in 9 of the 10 dataset--baseline pairs (\eg,
R0 on RAVDESS rises from 5.33 to 6.69; the exception is R2 on CREMA-D, from 2.47 to 2.23). Our event recall drops from
0.793 to 0.767, as do those of the baselines (A2F+LDN: 0.830 to 0.801), and the ranking is unchanged. Reporting the
validation numbers therefore does not change our conclusions.

\section{Evaluation Protocol}
\label{sec:supp_eval}

\paragraph{Event matching.} For each clip and family $f$, let $\mathcal{Q}_f=\{q_j\}$ be the specified events of the
input notes and $\mathcal{R}_f=\{r_i\}$ the re-encoded events, each occupying a frame interval $[a,b)$. With
\begin{equation}
\mathrm{IoU}(r,q)=\frac{\left|[a_r,b_r)\cap[a_q,b_q)\right|}{\left|[a_r,b_r)\cup[a_q,b_q)\right|},
\end{equation}
the candidate pairs are $\{(i,j):\mathrm{IoU}(r_i,q_j)\ge 0.5\}$. Pairs are taken in decreasing IoU (ties broken by
the index of the specified event, then of the re-encoded event) and accepted if neither end has been used, giving a
one-to-one matching $\mathcal{M}_f$; events of different families are never matched. Event recall and event intensity
agreement are
\begin{equation}
\begin{gathered}
\mathrm{Recall}=\frac{\sum\sum_f|\mathcal{M}_f|}{\sum\sum_f|\mathcal{Q}_f|},\\
\mathrm{IntAgr}=\frac{\sum\sum_f\sum_{(i,j)\in\mathcal{M}_f}\mathbf{1}[\ell(r_i)=\ell(q_j)]}{\sum\sum_f|\mathcal{M}_f|},
\end{gathered}
\end{equation}
where the outer sums run over clips (and seeds, counted together) and $\ell(\cdot)$ is the unsigned level; the
results are averaged within each dataset and then across datasets. The vision--language reader
(\cref{sec:supp_vlm}) uses the same matching.

\paragraph{Other execution metrics.} State level agreement is the fraction of frames with a specified state on which
the level is recovered; amplitude error is the absolute difference between the recovered and the specified
amplitude, divided by the specified amplitude and averaged over matched events. %

\paragraph{TAS.} TAS16 maps the re-encoded and the input notes to the 16 labels of Timeline Control~\cite{ma2025timeline} (8 regions, one
class per region per frame) and computes a per-frame F1 from per-class TP/FP/FN; macro averages within regions and
then across regions, and micro pools all classes. TAS45 does the same with the 45 labels of TL. Every format is
relabeled by the same encoder, and intensity is measured separately by event intensity agreement. Our TAS values are
not comparable with those reported by Timeline Control, since the label mapping, the encoder, the data and the sparsity of the director notes all differ. %

\paragraph{Seam.} Seam measures smoothness at the boundaries of the constraint mask. For each group of channels that the notes always constrain together (\eg, the six head-rotation channels) and each frame $t$ where the mask switches ($M_{t-1}\neq M_t$), we take the second difference (acceleration) over the six frames $t-3,\dots,t+2$, with each channel divided by its standard deviation, and average its RMS over the channels of the group. Seam is $|\log(\bar a/\bar a_{\text{real}})|$, where $\bar a$ is this mean over all boundaries and $\bar a_{\text{real}}$ is the same quantity for real motion at the same positions; 0 means as smooth as real motion. Boundaries whose frames $t-4,\dots,t+4$ are not all inside the clip and valid are skipped. %

\paragraph{FMD.} We use velocities rather than positions so that FMD reflects the dynamics rather than a person's
static expression, and compute it only on unconstrained entries so that only what the model completes is scored.
Velocities are cut into 30-frame windows and reduced by a PCA fitted on the training set before computing the
Fr\'echet distance~\cite{heusel2017fid}. \Cref{tab:supp_perturb} validates the metric with graded perturbations of real motion: FMD grows
monotonically with temporal smoothing, amplitude damping, additive noise and blink removal. It is nearly
insensitive to small high-frequency noise and to time shifts, which preserve the distribution of velocity windows;
seam and the W1 velocity distance (per channel group, the Wasserstein-1 distance between the distributions of
$|v|$, normalized by the training mean $|v|$) cover these cases.

\begin{table*}[t]
\centering
\footnotesize
\setlength{\tabcolsep}{3pt}
\begin{tabular}{@{}lccc@{}}
\toprule
Perturbation of real motion & FMD & W1 & Seam \\
\midrule
None (real) & 0 & 0 & 0 \\
Noise, 1 / 2 / 4 & 0.008 / 0.06 / 0.57 & 1.7 / 4.4 / 10.4 & 1.6 / 3.3 / 6.1 \\
Smoothing, 5 / 15 / 45 & 1.61 / 5.36 / 7.52 & 1.1 / 2.4 / 3.2 & 1.1 / 0.3 / 1.0 \\
Damping to 75 / 50 / 25\% & 0.54 / 2.13 / 4.60 & 0.9 / 1.7 / 2.6 & 0.4 / 1.1 / 1.8 \\
Time shift, 10 / 30 / 90 & 0.62 / 0.72 / 0.74 & 0.4 / 0.5 / 0.5 & 4.8 / 4.5 / 5.0 \\
Blinks removed & 1.52 & 0.6 & 0.2 \\
\textit{Initial generator (G0)} & \textit{1} & \textit{1} & \textit{1} \\
\bottomrule
\end{tabular}
\caption{Graded perturbations of real motion (validation set, macro over 3 datasets), relative to G0, an initial
version of our generator (= 1), the starting point of the configuration selection (LDN + script, no training variant and no blink masking); lower is closer to real motion. The perturbations were scored before the final configuration was fixed, so the table only shows how sensitive each metric is to each perturbation and is not compared with other tables; the final model has an FMD of 0.98 relative to G0 (standard notes). Only unconstrained entries are perturbed.
Noise $k$: i.i.d.\ Gaussian noise per frame with standard deviation $0.1k$ times that of the channel over the clip;
smoothing $w$: centred moving average over $w$ frames; damping: towards the neutral value (all channels except head
rotation); time shift $k$: the signal taken $k$ frames (30\,Hz) later; blinks removed: blink channels set to the
neutral value.}
\label{tab:supp_perturb}
\end{table*}

\paragraph{Leakage.} We specify a single instruction (target family $f^\ast$, span $[t_a,t_d)$) with everything else
empty, and generate once; with the same script, clip header and noise but nothing specified, we generate again as a
control. For every timeline $k$ outside $f^\ast$,
\begin{equation}
D_k=\frac{\left|\bar s_k(\hat X_{e})-\bar s_k(\hat X_{\varnothing})\right|}{e_{0,k}},\qquad
\mathcal{L}_e=\frac{1}{|\mathcal{K}|}\sum_{k\in\mathcal{K}}D_k,
\end{equation}
where $\bar s_k$ is the mean of the state curve of timeline $k$ over $[t_a,t_d)$ (for event instructions, the
unsmoothed value, to keep short changes), and $e_{0,k}$ is the lowest level boundary of $k$, the minimum change the
encoder can detect; $D_k=1$ means one minimum detectable change. A fixed boundary is used instead of each clip's
standard deviation, which is unstable on short spans. We report $\mathcal{L}_e$ and the fraction with $D_k\ge1$; the
noise floor applies the same computation to two generations with nothing specified and different noise seeds.

\paragraph{Statistics.} Mean metrics use a speaker-level bootstrap (95\% intervals), so that clips of the same speaker
are not treated as independent. FMD is not bootstrapped: resampling duplicates clips in the real reference set and
raises its own FMD by 22--55\%. FMD comparisons therefore use a paired permutation test that swaps the outputs of two
methods per clip with probability 1/2 (5000 permutations, two-sided), keeping the real data fixed. The main paper
reports no significance tests except in the user study; they are collected here.

\section{Additional Results}
\label{sec:supp_results}

\subsection{Seeds, Absolute FMD and Statistical Tests}

Over the 3 seeds of our generator, Tab.~1 gives relative FMD $1.00\pm0.007$ (0.992 / 1.007 / 1.001), seam
$0.089\pm0.003$, event recall $0.744\pm0.004$, state execution $0.326\pm0.002$ and tendency execution
$0.615\pm0.006$. The absolute velocity FMD of our model is 4.60 on TalkVid, 11.33 on RAVDESS and 10.27 on CREMA-D.
The baselines are deterministic and have no seeds. On standard notes, the speaker-bootstrap differences of our model
are $-0.037$ $[-0.068,-0.008]$ (event), $+0.279$ $[0.263, 0.292]$ (state) and $+0.101$ $[0.082, 0.119]$ (tendency)
against R1, and $-0.073$ $[-0.100,-0.045]$, $+0.217$ $[0.194, 0.235]$ and $+0.106$ $[0.089, 0.125]$ against A2F+LDN.
\Cref{tab:supp_tab2} gives the per-seed standard deviations of Tab.~2 and the LDN differences with their intervals.
On TAS16 macro, LDN and TL differ by 0.003, far below the seed standard deviation (0.03--0.04).

\begin{table*}[t]
\centering
\footnotesize
\setlength{\tabcolsep}{8pt}
\begin{tabular}{@{}lcccc@{}}
\toprule
 & TC16 & TL & LDN & LDN $-$ TL \\
\midrule
TAS16 macro & 0.441$\pm$0.024 & 0.468$\pm$0.031 & 0.471$\pm$0.037 & $+$0.003 [$-$0.008, $+$0.035] \\
TAS16 micro & 0.566$\pm$0.006 & 0.564$\pm$0.004 & 0.590$\pm$0.005 & $+$0.026 [$+$0.015, $+$0.038] \\
TAS45 macro & 0.245$\pm$0.007 & 0.321$\pm$0.014 & 0.361$\pm$0.020 & $+$0.040 [$+$0.030, $+$0.060] \\
Event intensity agreement & 0.492$\pm$0.022 & 0.536$\pm$0.022 & 0.823$\pm$0.016 & $+$0.287 [$+$0.254, $+$0.321] \\
Amplitude error & 0.373$\pm$0.016 & 0.383$\pm$0.003 & 0.124$\pm$0.007 & $-$0.26 [$-$0.29, $-$0.23] \\
State level agreement & 0.228$\pm$0.008 & 0.241$\pm$0.006 & 0.277$\pm$0.005 & $+$0.035 [$+$0.020, $+$0.051] \\
\bottomrule
\end{tabular}
\caption{Tab.~2 with the standard deviation over 3 seeds and the speaker-bootstrap 95\% interval of the
LDN$-$TL difference. LDN$-$TC16: $+$0.030 [$+$0.015, $+$0.084], $+$0.024 [$+$0.013, $+$0.036], $+$0.116 [$+$0.101,
$+$0.133], $+$0.331 [$+$0.288, $+$0.370], $-$0.25 [$-$0.28, $-$0.22], $+$0.049 [$+$0.030, $+$0.065] (same row order).}
\label{tab:supp_tab2}
\end{table*}

The all-frame FMD of TL and TC16 relative to LDN (1.06 and 1.08, Sec.~4.4) has permutation $p=0.23$ and $p=0.09$;
per dataset, the ratio LDN/TL is 1.064 on TalkVid, 0.937 on RAVDESS and 0.855 on CREMA-D.

\subsection{Held-Out Test Speakers}
\label{sec:supp_test}

\Cref{tab:supp_test} repeats Tab.~1 on the test speakers of RAVDESS and CREMA-D (1300 clips, 17 speakers never used
for training, calibration or model selection), with the encoder constants unchanged. TalkVid test speakers are not included because
no word timing was transcribed for them. Every comparison keeps its direction; the state execution margin over
interpolation is $+0.32$ $[0.29, 0.34]$ (validation set: $+0.32$).

\begin{table}[!htb]
\centering
\footnotesize
\setlength{\tabcolsep}{2.5pt}
\resizebox{\linewidth}{!}{%
\begin{tabular}{@{}lcccc@{}}
\toprule
Method & FMD $\downarrow$ & Event $\uparrow$ & State $\uparrow$ & Tendency $\uparrow$ \\
\midrule
R0 Neutral & 7.52 & 0.904 \textit{(0.900)} & 0.054 \textit{(0.054)} & 0.673 \textit{(0.638)} \\
R1 Interpolation & 7.69 & 0.843 \textit{(0.864)} & 0.055 \textit{(0.055)} & 0.672 \textit{(0.637)} \\
R2 Retrieval & 1.87 & 0.587 \textit{(0.601)} & 0.186 \textit{(0.159)} & 0.601 \textit{(0.580)} \\
A2F & 8.76 & 0.089 \textit{(0.084)} & 0.095 \textit{(0.082)} & 0.635 \textit{(0.614)} \\
A2F+LDN & 6.17 & 0.801 \textit{(0.830)} & 0.146 \textit{(0.135)} & 0.649 \textit{(0.633)} \\
Ours & 1 & 0.767 \textit{(0.793)} & 0.376 \textit{(0.369)} & 0.716 \textit{(0.692)} \\
\textit{Human} & --- & 0.999 \textit{(0.999)} & 0.908 \textit{(0.883)} & 0.999 \textit{(0.997)} \\
\bottomrule
\end{tabular}}
\caption{Test split (RAVDESS + CREMA-D, standard notes, macro over the two datasets). FMD on unconstrained entries,
Ours = 1. In parentheses: the validation set on the same two datasets.}
\label{tab:supp_test}
\end{table}

\subsection{Full and Lite Notes}

\Cref{tab:supp_density_rows} gives the main comparison at the three densities. The conclusions are the same at all
densities. On lite notes, our seam relative to the initial generator (G0, \cref{tab:supp_perturb}) is high (8.4); we have not analysed the cause.

\begin{table}[!htb]
\centering
\footnotesize
\setlength{\tabcolsep}{2pt}
\resizebox{\linewidth}{!}{%
\begin{tabular}{@{}lcccccccc@{}}
\toprule
 & \multicolumn{4}{c}{Full notes} & \multicolumn{4}{c}{Lite notes} \\
\cmidrule(lr){2-5}\cmidrule(l){6-9}
Method & FMD & Event & State & Tend. & FMD & Event & State & Tend. \\
\midrule
R0 Neutral & 6.15 & 0.907 & 0.146 & 0.723 & 9.06 & 0.757 & 0.012 & 0.497 \\
R1 Interp. & 8.42 & 0.524 & 0.273 & 0.647 & 9.08 & 0.745 & 0.012 & 0.498 \\
R2 Retrieval & 2.56 & 0.469 & 0.332 & 0.545 & 1.81 & 0.543 & 0.102 & 0.462 \\
A2F & 13.72 & 0.052 & 0.066 & 0.490 & 8.45 & 0.079 & 0.070 & 0.490 \\
A2F+LDN & 6.67 & 0.849 & 0.215 & 0.711 & 7.78 & 0.718 & 0.082 & 0.492 \\
Ours & 1 & 0.697 & 0.430 & 0.684 & 1 & 0.666 & 0.352 & 0.589 \\
\textit{Human} & --- & 0.996 & 1.000 & 0.991 & --- & 0.996 & 0.772 & 0.991 \\
\bottomrule
\end{tabular}}
\caption{Main comparison on full and lite notes (macro over the three datasets; FMD on unconstrained entries,
relative to Ours in each density). Event, state and tendency execution as in Tab.~1.}
\label{tab:supp_density_rows}
\end{table}

\subsection{Effect of the Projection}

Disabling the projection at every DDIM step and after smoothing (same model and noise) lowers event recall from
0.744 to 0.676 ($-0.068$ $[-0.086,-0.050]$), while state execution (0.326 vs.\ 0.323), tendency execution (0.615 vs.\
0.614), FMD (1.01 relative to the projected output) and seam (0.94) change little. Without projection the constrained
entries leave the tolerance by up to 0.51 (blendshape) and $2.39^\circ$ (head rotation). On the 299 clips of Tab.~2,
the output of Tab.~1 has event intensity agreement and amplitude error of 0.832 / 12.1\% with projection (Tab.~2, a
separate run: 0.823 / 12.4\%) and 0.668 / 20.9\% without, against 0.536 / 38.3\% for TL; the difference in amplitude error between our unprojected output and TL is $-0.174$
$[-0.206,-0.141]$.

\subsection{Layered versus Single-Layer Notes}
\label{sec:supp_flat}

A natural question is how a single-layer format would compare with LDN if it carried the same information. We
argue that such a fair single-layer format does not exist.

\paragraph{A fair single layer must express superposition.} Facial motion mixes components of different duration:
expressions that last a second or more, and rapid rises and falls on top of them, such as a smile that briefly
deepens while it is held, or a nod while the head stays lowered. Such superposition is common: on the validation
set, 18\% of the events in the standard notes and 26\% in the full notes lie on a state of the same channel. To
carry the same information as LDN, a single-layer format must express on one track both how high the sustained
part is and how fast and how strongly the motion on top of it rises and falls.

\paragraph{Extracting superposition is layered extraction.} To obtain such a representation from tracked curves, the
slow part must be separated from the rapid motion, and the amplitude of the rapid motion must be measured above the
slow part rather than from the neutral value. This separation is exactly what our encoder does: it extracts states by
smoothing and detects events on the residual after the states are removed, so the amplitude of an event is its
deviation above the state (Sec.~3.3). A single-layer format that skips this step has only labels, like TL and TC16:
a rapid motion on top of a sustained one is either merged into it or cuts it into pieces, and its intensity cannot
be expressed. Such a format carries less information than LDN and is therefore not a fair comparison. In Tab.~2,
event intensity agreement is 0.54 and 0.49 for TL and TC16 against 0.82 for LDN, and amplitude error is 0.38 and 0.37
against 0.12.

\paragraph{Flattening layered notes keeps the layers.} Conversely, one can extract layered notes first and then
flatten them onto one timeline. We build such a format, FL: the state and the event of the same channel and direction
are merged into one timeline by taking the larger value per frame (27 timelines), and all the information of LDN is kept
(\cref{tab:supp_flat_def}). Every quantity it keeps is defined by the layers: the intensity of an event is its
amplitude above the state, divided by the representative event value of that timeline; the onset and offset come from the
four time points of each state or event; every entry has its own specified flag; and the tendencies enter as constant
timelines. The FL generator uses the same network, data, training and 3 seeds as LDN, and keeps the constrained entries
and the projection. FL only rearranges the layered content and remains layered.

A single-layer format is thus either not layered and carries less information than LDN (TL, TC16), or carries the
same information and is itself the result of layering (FL); a format with the same information but without layering
cannot be constructed. The question of layered versus single-layer notes comes down to whether motions of different
duration are fitted separately.

\paragraph{Experimental check.} The argument predicts that FL performs like LDN and differs only where flattening
itself loses something. The results agree.
\begin{itemize}[leftmargin=*,itemsep=0pt,topsep=2pt]
\item \emph{Same execution} (\cref{tab:supp_flat_gen}). With the real notes as input, FL and LDN differ by at most
0.015 on every metric of Tab.~2; the only interval that excludes zero is TAS45 macro, where FL is 0.014 higher. The
all-frame FMD ratio LDN\,/\,FL is 0.98 $[0.97, 1.03]$. Both are far better than TL.
\item \emph{Cost of flattening: masking at overlaps} (\cref{tab:supp_flat_overlap}). Taking the larger value per frame
hides an event that is weaker than the state beneath it. On full notes, LDN recalls 0.035 $[0.019, 0.050]$ more of the
overlapping events but has 0.031 $[0.014, 0.050]$ higher amplitude error there; on standard notes the differences are
not significant.
\item \emph{Cost of flattening: implicit layers are harder to author} (\cref{tab:supp_flat_llm}). When a language model
authors the single-layer text (FL text, with the same clips as Sec.~4.6), the layer of each entry is only implied by the
shape of its fields. With Opus the two formats are close: the strict validity is 0.963 for LDN and 0.920 for FL text, and
on the 292 clips valid in both formats only event intensity agreement differs ($+0.034$ $[0.008, 0.060]$ for LDN).
With Haiku the validity after repair is 0.936 for LDN and 0.505 for FL text; 76 of the 148 invalid FL text notes make the
same error, a head rotation given a vector value but no kind, that is, a wrong layer. On the 146 clips valid in
both, LDN is higher by 0.062 $[0.035, 0.086]$ on TAS45 micro and by 0.058 $[0.022, 0.090]$ on state level agreement.
\end{itemize}
For naturalness, with Opus as the author, the all-frame FMD of the LDN pipeline is 1.08 $[1.02, 1.10]$ times that of
the FL text pipeline. The difference comes from the generator, not from the authored format: with the FL generator,
LDN-authored and FL-text-authored notes give a ratio of 1.00 $[0.97, 1.06]$, and with the real notes as input the two
generators give 0.98. With Haiku, half of the FL text notes are invalid and count as empty notes, so its FMD is not
comparable.

\begin{table}[!htb]
\centering
\footnotesize
\setlength{\tabcolsep}{2pt}
\begin{tabular}{@{}lp{0.74\linewidth}@{}}
\toprule
Part & FL (68 input columns) \\
\midrule
Timelines & 45 LDN labels without the state / event prefix, merged into 27 timelines (larger value per frame) \\
Intensity & $|v|$ divided by the training-set value of the moderate level of that timeline; level value if no value is given; clipped to $[0,3]$ \\
Shape & intensity $\times$ onset / plateau / offset from the four time points \\
Specified & one 0/1 column per timeline (1 on frames with an entry) \\
Tendencies & 7 constant timelines ($-1$ to $1$) and 7 specified flags \\
Execution & constrained entries and projection, as in LDN \\
\bottomrule
\end{tabular}
\caption{Construction of FL from the same notes as LDN.}
\label{tab:supp_flat_def}
\end{table}

\begin{table}[!htb]
\centering
\footnotesize
\setlength{\tabcolsep}{1.6pt}
\resizebox{\linewidth}{!}{%
\begin{tabular}{@{}lcccc@{}}
\toprule
 & TL & FL & LDN & LDN $-$ FL \\
\midrule
TAS16 micro & 0.565 & 0.589 & 0.591 & $+$0.002 [$-$0.005, $+$0.009] \\
TAS45 micro & 0.480 & 0.517 & 0.514 & $-$0.003 [$-$0.009, $+$0.005] \\
TAS45 macro & 0.306 & 0.360 & 0.346 & $-$0.014 [$-$0.024, $-$0.001] \\
Event intensity agreement & 0.501 & 0.830 & 0.832 & $+$0.002 [$-$0.017, $+$0.020] \\
Amplitude error & 0.427 & 0.120 & 0.124 & $+$0.004 [$-$0.002, $+$0.011] \\
State level agreement & 0.253 & 0.280 & 0.272 & $-$0.008 [$-$0.019, $+$0.002] \\
\bottomrule
\end{tabular}}
\caption{FL against LDN and TL with the real standard notes (299 validation clips $\times$ 3 seeds, each format with
its own generator, same metrics as Tab.~2). This run uses one sample per seed, so LDN and TL differ slightly from
Tab.~2. Intervals: speaker bootstrap, 95\%.}
\label{tab:supp_flat_gen}
\end{table}

\begin{table}[!htb]
\centering
\footnotesize
\setlength{\tabcolsep}{3pt}
\resizebox{\linewidth}{!}{%
\begin{tabular}{@{}lccc@{}}
\toprule
 & LDN & FL & LDN $-$ FL [95\% CI] \\
\midrule
\multicolumn{4}{@{}l}{\textit{Overlap, standard notes (417 of 2376 events)}} \\
Event recall $\uparrow$ & 0.671 & 0.705 & $-$0.034 [$-$0.072, $+$0.014] \\
Event intensity agreement $\uparrow$ & 0.832 & 0.779 & $+$0.053 [$-$0.004, $+$0.106] \\
Amplitude error $\downarrow$ & 0.186 & 0.168 & $+$0.019 [$-$0.000, $+$0.038] \\
State level agreement $\uparrow$ & 0.248 & 0.263 & $-$0.015 [$-$0.043, $+$0.011] \\
\midrule
\multicolumn{4}{@{}l}{\textit{Overlap, full notes (3498 of 13299 events)}} \\
Event recall $\uparrow$ & 0.670 & 0.636 & $+$0.035 [$+$0.019, $+$0.050] \\
Event intensity agreement $\uparrow$ & 0.606 & 0.607 & $-$0.001 [$-$0.027, $+$0.021] \\
Amplitude error $\downarrow$ & 0.303 & 0.272 & $+$0.031 [$+$0.014, $+$0.050] \\
State level agreement $\uparrow$ & 0.481 & 0.489 & $-$0.008 [$-$0.024, $+$0.010] \\
\midrule
\multicolumn{4}{@{}l}{\textit{Elsewhere, standard notes}} \\
Event recall $\uparrow$ & 0.726 & 0.749 & $-$0.023 [$-$0.052, $+$0.001] \\
Event intensity agreement $\uparrow$ & 0.832 & 0.840 & $-$0.008 [$-$0.029, $+$0.011] \\
Amplitude error $\downarrow$ & 0.112 & 0.110 & $+$0.002 [$-$0.005, $+$0.009] \\
State level agreement $\uparrow$ & 0.276 & 0.283 & $-$0.007 [$-$0.018, $+$0.003] \\
\midrule
\multicolumn{4}{@{}l}{\textit{Elsewhere, full notes}} \\
Event recall $\uparrow$ & 0.699 & 0.711 & $-$0.011 [$-$0.021, $-$0.001] \\
Event intensity agreement $\uparrow$ & 0.758 & 0.758 & $+$0.001 [$-$0.013, $+$0.013] \\
Amplitude error $\downarrow$ & 0.133 & 0.133 & $+$0.000 [$-$0.004, $+$0.005] \\
State level agreement $\uparrow$ & 0.320 & 0.323 & $-$0.003 [$-$0.019, $+$0.015] \\
\bottomrule
\end{tabular}}
\caption{LDN and FL where a state and an event overlap on the same channel, and elsewhere (real notes, each
format with its own generator). Intervals: speaker bootstrap, 95\%.}
\label{tab:supp_flat_overlap}
\end{table}

\begin{table}[!htb]
\centering
\footnotesize
\setlength{\tabcolsep}{2pt}
\resizebox{\linewidth}{!}{%
\begin{tabular}{@{}lcc@{}}
\toprule
 & Opus & Haiku \\
\midrule
Validity after repair, LDN / FL text & 0.990 / 0.987 & 0.936 / 0.505 \\
Strict validity, LDN / FL text & 0.963 / 0.920 & 0.656 / 0.445 \\
Clips valid in both & 292 & 146 \\
TAS45 micro, LDN $-$ FL text & $+$0.003 [$-$0.007, $+$0.014] & $+$0.062 [$+$0.035, $+$0.086] \\
Event int.\ agr., LDN $-$ FL text & $+$0.034 [$+$0.008, $+$0.060] & $+$0.070 [$-$0.005, $+$0.131] \\
State level agr., LDN $-$ FL text & $+$0.001 [$-$0.017, $+$0.020] & $+$0.058 [$+$0.022, $+$0.090] \\
\bottomrule
\end{tabular}}
\caption{Language models authoring LDN and the single-layer text FL text, the input of the FL generator, (299 validation clips, each format with its own
generator, 3 seeds). Execution is measured against the notes the model authored, on the clips valid in both formats.}
\label{tab:supp_flat_llm}
\end{table}

\subsection{\texorpdfstring{Encoder--Decoder Fidelity and Recalibration}{Encoder--Decoder Fidelity and Recalibration}}

\Cref{tab:supp_closure} gives the per-family error of decoding the full notes, the basis of Sec.~4.2. Each entry is
the mean absolute error between the decoded (or generated) and the real curves over all channels and frames of the
family. The neutral-value column is a scale, not a method; errors are compared only within a row, since their size
also depends on how much a family moves. Passing the full notes through the generator raises the error slightly,
because the generator completes motion on the unconstrained entries, so the curve no longer follows the real one
frame by frame. The reconstruction is close but not exact; the missed events are mainly brow raises and head rotations
(\cref{tab:supp_family}). \Cref{tab:supp_reencode} gives the re-encoding results at the three densities, and
\cref{tab:supp_refit} the recalibration experiment. Direction agreement is the fraction of matched events with the
same direction (\eg, both turning left).

\begin{table}[!htb]
\centering
\footnotesize
\setlength{\tabcolsep}{3pt}
\begin{tabular}{@{}lccc@{}}
\toprule
Family & Neutral only & Decode & Decode + generator \\
\midrule
Eyelids & 0.053 & 0.026 & 0.027 \\
Squint & 0.100 & 0.050 & 0.057 \\
Brow raise & 0.081 & 0.043 & 0.048 \\
Frown & 0.058 & 0.023 & 0.032 \\
Smile & 0.147 & 0.057 & 0.068 \\
Mouth tension & 0.048 & 0.030 & 0.030 \\
Head rotation ($^\circ$) & 1.64 & 0.94 & 1.38 \\
Head translation & 0.377 & 0.304 & 0.407 \\
Gaze & 0.016 & 0.009 & 0.013 \\
\bottomrule
\end{tabular}
\caption{Reconstruction from full notes (validation set, macro over datasets): mean absolute error against the real curves.
Blendshape families on the $[0,1]$ scale; head translation in tracker units; gaze as the iris-centre offset divided by the eye width (dimensionless).}
\label{tab:supp_closure}
\end{table}

\begin{table}[!htb]
\centering
\footnotesize
\setlength{\tabcolsep}{2.5pt}
\resizebox{\linewidth}{!}{%
\begin{tabular}{@{}llcccc@{}}
\toprule
Notes & Re-encoded from & Event F1 & Recall & State agr. & Tend.\ agr. \\
\midrule
Full & decoded & 0.882 & 0.836 & 0.889 & 0.707 \\
 & generated & 0.664 & 0.699 & 0.840 & 0.680 \\
Standard & decoded & 0.261 & 0.837 & 0.847 & 0.516 \\
 & generated & 0.301 & 0.738 & 0.811 & 0.609 \\
Lite & decoded & 0.054 & 0.741 & 0.853 & 0.501 \\
 & generated & 0.209 & 0.664 & 0.808 & 0.592 \\
\bottomrule
\end{tabular}}
\caption{Re-encoding decoded and generated curves. Event F1, state and tendency agreement are against the full
notes of the real clip; recall is against the specified events. (Generated: an earlier evaluation run of the
final model, without the projection after smoothing.)}
\label{tab:supp_reencode}
\end{table}

\begin{table}[!htb]
\centering
\footnotesize
\setlength{\tabcolsep}{2pt}
\resizebox{\linewidth}{!}{%
\begin{tabular}{@{}llccccc@{}}
\toprule
Data & Calib. & Ev.\ micro & Ev.\ macro & Dir. & State & Tend. \\
\midrule
TalkVid & T vs.\ R+C & 0.937 & 0.936 & 0.976 & 0.899 & 0.235 \\
 & $T_a$ vs.\ $T_b$ & 0.966 & 0.965 & 0.987 & 0.964 & 0.876 \\
RAVDESS & R vs.\ T & 0.954 & 0.951 & 0.987 & 0.948 & 0.417 \\
 & R vs.\ C & 0.968 & 0.967 & 0.993 & 0.979 & 0.738 \\
 & $R_a$ vs.\ $R_b$ & 0.958 & 0.951 & 0.997 & 0.968 & 0.698 \\
CREMA-D & C vs.\ T & 0.969 & 0.972 & 0.992 & 0.965 & 0.512 \\
 & C vs.\ R & 0.980 & 0.978 & 0.995 & 0.983 & 0.810 \\
 & $C_a$ vs.\ $C_b$ & 0.982 & 0.982 & 0.998 & 0.986 & 0.950 \\
\bottomrule
\end{tabular}}
\caption{Recalibrating the encoder constants. Each row encodes the validation clips of one dataset with two sets of
constants calibrated on different data (T, R, C: TalkVid, RAVDESS, CREMA-D training sets; $X_a$ / $X_b$: two
speaker-disjoint halves of one dataset, the noise floor) and compares the notes. Dir.: direction agreement of
matched events.}
\label{tab:supp_refit}
\end{table}
When the TalkVid constants used in the paper are applied to RAVDESS and CREMA-D (R vs.\ T, C vs.\ T), the tendency agreement is only 0.42--0.51 (noise floor 0.70--0.95), because tendency levels are quintiles of the population used for calibration (Sec.~3.2). %

\subsection{Event Recall by Family}

\begin{table}[!htb]
\centering
\footnotesize
\setlength{\tabcolsep}{3pt}
\begin{tabular}{@{}lcccc@{}}
\toprule
Family & R0 Neutral & R1 Interp. & A2F+LDN & Ours \\
\midrule
Blink & 1.00 & 0.94 & 1.00 & 0.93 \\
Smile & 0.94 & 0.91 & 0.87 & 0.90 \\
Gaze & 0.90 & 0.76 & 0.90 & 0.77 \\
Mouth tension & 0.94 & 0.89 & 0.70 & 0.84 \\
Frown & 0.90 & 0.88 & 0.90 & 0.76 \\
Brow raise & 0.82 & 0.81 & 0.86 & 0.70 \\
Squint & 0.89 & 0.87 & 0.81 & 0.68 \\
Head rotation & 0.75 & 0.64 & 0.75 & 0.55 \\
\bottomrule
\end{tabular}
\caption{Event recall per family (standard notes, counts pooled over the three datasets, so the values differ
slightly from the macro averages of Tab.~1).}
\label{tab:supp_family}
\end{table}

By family (\cref{tab:supp_family}), we are within 0.01 of interpolation on blinks, smile and gaze, and 0.09--0.19
lower on squint, brow raise, frown and head rotation. %
Classifying each specified event as recovered, only partly recovered (a short segment), absorbed by a longer event or
missed, the shares are 0.891 / 0.075 / 0.000 / 0.034 for R0, 0.849 / 0.102 / 0.004 / 0.045 for decoding alone,
0.828 / 0.103 / 0.021 / 0.049 for R1 and 0.764 / 0.119 / 0.029 / 0.088 for ours (pooled). Partial recovery occurs for
every method, including decoding alone (the recovered segment is about 35\% of the specified length, with its centre
about 9 frames off), so it comes from decoding and re-encoding themselves; R0 is high because the motion outside the
events is still and has no states. We have not identified the cause of the fully missed events.

\paragraph{Events in unspecified spans.} Re-encoded events per minute are 141 for ours, 166 for real motion, 30 for
R1 and 76 for A2F+LDN; blinks 23.5 vs.\ 27.5 per minute and head motions 18.2 vs.\ 23.6 (ours vs.\ real). Of the
events re-encoded from our output, 16\% match a specified event, 17\% are unspecified but also present in the real
motion, and 67\% are in neither. To suppress motion, a tendency can be specified (\eg, a very low eyeblink frequency).

\subsection{Intensity, Tendencies and Leakage}
\setcounter{topnumber}{1}\setcounter{bottomnumber}{0}

\begin{figure*}[t]
\centering
\includegraphics[width=\textwidth]{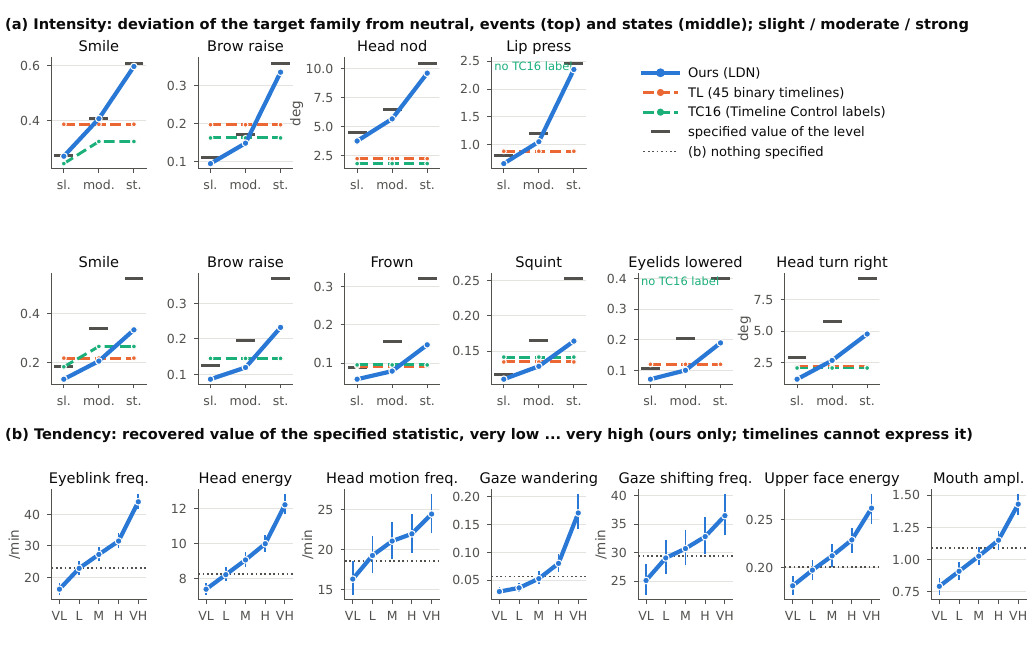}
\caption{Full version of Fig.~6 (299 validation clips $\times$ 3 seeds). \textbf{(a)} Every intensity instruction at the
slight, moderate and strong levels; short bars: the specified value of the level. \textbf{(b)} Every tendency from very
low to very high; dotted: nothing specified; error bars: 95\% speaker-bootstrap intervals. Numbers in
\cref{tab:supp_intensity,tab:supp_tendency}.}
\label{fig:supp_control}
\end{figure*}

\Cref{fig:supp_control} and \cref{tab:supp_intensity} give the full intensity experiment of Fig.~6 (a, b). The value specified with each
instruction is the training-set median of that level, with the same duration and position for all levels. We
compare the absolute deviation from neutral on the plateau with the specified value; subtracting the output with
nothing specified would remove motion the model performs anyway and underestimate execution. The response of LDN
rises monotonically with the level for 99.5--100\% of the clips (per-clip rank correlation 0.92--1.00). Since
timeline labels carry no intensity, the three levels give the same output for them, except that TC16 distinguishes two
levels of smile. %

\begin{table*}[t]
\centering
\footnotesize
\setlength{\tabcolsep}{4pt}
\begin{tabular}{@{}lccccccc@{}}
\toprule
Instruction & Specified (sl./mod./str.) & LDN output (sl./mod./str.) & LDN / specified & None & TL & TC16 \\
\midrule
State: smile & 0.18 / 0.34 / 0.54 & 0.13 / 0.21 / 0.33 & 0.71 / 0.60 / 0.61 & 0.12 & 0.22 & 0.18 / 0.27 / 0.27 \\
State: brow raise & 0.13 / 0.19 / 0.37 & 0.09 / 0.12 / 0.23 & 0.69 / 0.62 / 0.63 & 0.08 & 0.15 & 0.15 \\
State: frown & 0.09 / 0.16 / 0.32 & 0.06 / 0.08 / 0.15 & 0.66 / 0.51 / 0.46 & 0.07 & 0.09 & 0.10 \\
State: squint & 0.12 / 0.17 / 0.25 & 0.11 / 0.13 / 0.16 & 0.94 / 0.78 / 0.65 & 0.10 & 0.14 & 0.14 \\
State: eyelids lowered & 0.11 / 0.20 / 0.40 & 0.07 / 0.10 / 0.19 & 0.68 / 0.50 / 0.48 & 0.07 & 0.12 & 0.07$^\ast$ \\
State: head turn ($^\circ$) & 2.9 / 5.8 / 9.2 & 1.2 / 2.7 / 4.8 & 0.42 / 0.47 / 0.52 & $-$0.1 & 2.2 & 2.1 \\
Event: smile & 0.28 / 0.41 / 0.61 & 0.27 / 0.41 / 0.60 & 0.99 / 1.00 / 0.98 & 0.15 & 0.39 & 0.25 / 0.33 / 0.33 \\
Event: brow raise & 0.11 / 0.17 / 0.36 & 0.09 / 0.15 / 0.34 & 0.86 / 0.87 / 0.93 & 0.08 & 0.20 & 0.16 \\
Event: nod ($^\circ$) & 4.5 / 6.5 / 10.5 & 3.7 / 5.7 / 9.6 & 0.83 / 0.87 / 0.92 & $-$0.6 & 2.2 & 1.8 \\
Event: lip press & 0.80 / 1.20 / 2.47 & 0.66 / 1.05 / 2.35 & 0.83 / 0.88 / 0.95 & 0.64 & 0.88 & 0.63$^\ast$ \\
\bottomrule
\end{tabular}
\caption{Intensity (299 validation clips $\times$ 3 seeds, macro over datasets): absolute deviation of the target
family from neutral on the plateau (state value) or at the apex (event), in the family's own units. None: nothing
specified. TL and TC16 receive the same input for all levels (one value), except TC16 smile (slight / moderate,
strong). $^\ast$TC16 has no label for lip press and lowered eyelids, so the instruction cannot be given and its
output equals its output with nothing specified.}
\label{tab:supp_intensity}
\end{table*}

\Cref{tab:supp_tendency} gives the tendency experiment of Fig.~6c: one tendency is specified from very low to very
high, with no value and nothing else, and the generated result is re-encoded to recover the actual value. The
exact agreement between the recovered level and the specified level is 0.23--0.30 (0.20 at chance); agreement within
one level is 0.50--0.66. %
The timeline formats have no labels for such whole-clip instructions. %

\begin{table*}[t]
\centering
\footnotesize
\setlength{\tabcolsep}{4pt}
\begin{tabular}{@{}lcccccccccc@{}}
\toprule
Tendency & None & VL & L & M & H & VH & Slope [95\% CI] & $\rho$ & Exact & $\pm1$ \\
\midrule
Eyeblink frequency (/min) & 22.95 & 16.34 & 22.94 & 27.32 & 31.55 & 43.94 & 27.6 [26.2, 29.0] & 0.970 & 0.297 & 0.628 \\
Head energy & 8.26 & 7.40 & 8.23 & 9.07 & 9.99 & 12.21 & 4.82 [4.58, 5.08] & 1.000 & 0.236 & 0.517 \\
Head motion frequency (/min) & 18.60 & 16.32 & 19.25 & 21.05 & 21.95 & 24.41 & 8.09 [7.13, 9.02] & 0.919 & 0.226 & 0.503 \\
Gaze wandering & 0.056 & 0.029 & 0.036 & 0.052 & 0.080 & 0.171 & 0.141 [0.118, 0.168] & 0.993 & 0.257 & 0.571 \\
Gaze shifting frequency (/min) & 29.45 & 25.12 & 29.08 & 30.73 & 32.83 & 36.50 & 11.4 [10.0, 12.6] & 0.932 & 0.235 & 0.555 \\
Upper face energy & 0.201 & 0.181 & 0.197 & 0.212 & 0.229 & 0.262 & 0.081 [0.072, 0.089] & 0.996 & 0.250 & 0.551 \\
Mouth amplitude & 1.090 & 0.793 & 0.910 & 1.027 & 1.150 & 1.429 & 0.636 [0.599, 0.673] & 1.000 & 0.278 & 0.656 \\
\bottomrule
\end{tabular}
\caption{Tendency control (validation set, 3 seeds pooled, macro over datasets): recovered value of the specified
statistic for each level (VL \dots{} VH) and with nothing specified. Slope: change from VL to VH;
$\rho$: rank correlation with the level; Exact / $\pm1$: fraction of recovered levels equal to / within one level of the
specified one (chance 0.20 / 0.52).}
\label{tab:supp_tendency}
\end{table*}

\paragraph{Leakage.} On the same experiments, we measure leakage (\cref{sec:supp_eval}; \cref{tab:supp_leak}).
When a single instruction is specified, the other families move by only 0.13--0.23 minimum detectable changes on average, less than 60\% of the random change when only the noise seed changes, and the fraction that the encoder can recover is only 1--6\%. The leakage of events is similar to that of the timeline formats. The leakage of states is slightly higher and mainly leaks into frown and smile; that is, when a state is specified, the expression of the whole face changes slightly. %

\begin{table*}[t]
\centering
\footnotesize
\setlength{\tabcolsep}{2.5pt}
\begin{tabular}{@{}lccccc@{}}
\toprule
Instr. & Ours: mean & Ours: $\ge$1 & Noise floor & TL & TC16 \\
\midrule
Events (4) & 0.13--0.21 & 1--3\% & 0.51--0.62 (14--19\%) & 0.09--0.18 & 0.09--0.14 \\
States (6) & 0.16--0.23 & 3--6\% & 0.34--0.41 (6--10\%) & 0.05--0.11 & 0.06--0.12 \\
\bottomrule
\end{tabular}
\caption{How much the other families are moved when only a single moderate-level instruction is specified (in units of the minimum detectable change; 1 means the encoder can just recover it); each entry is the min--max over instructions. ${\ge}1$ is the fraction of timelines with a change of at least 1; the noise floor is the same statistic with nothing specified and only the noise seed changed, and the values in parentheses give the fraction with ${\ge}1$ among them. TC16 does not include lip press and lowered eyelids, for which it has no labels.} %
\label{tab:supp_leak}
\end{table*}

\subsection{Stacking Layers and Density}

The director notes in Fig.~1 contain tendencies, states and events at the same time. We take the real standard notes
of each clip and keep only one, two or all three layers of them (\cref{tab:supp_layers}). When all three layers are
specified together, the execution of each layer is no worse than when it is specified alone: event recall rises from
0.686 to 0.746, state execution from 0.315 to 0.334 and tendency execution from 0.599 to 0.609. The three layers can
be stacked without interfering with one another. For naturalness, specifying events makes the motion closer to real
human motion (all-frame FMD drops from 19.3 to 16.7). %

For density, we prune the full notes by salience to $K$ events per minute, with the number of state segments decreasing with $K$ (max$(1,K/5)$ per minute), from empty to full notes (\cref{tab:supp_density}). %
Event recall stays at 0.70--0.75 at every density, and specifying more brings the result closer to real motion: the
all-frame FMD drops by 42\% from empty to full notes. Even full notes constrain only 13\% of the entries; the model
completes the rest. %

\begin{table}[tb]
\centering
\footnotesize
\setlength{\tabcolsep}{3pt}
\begin{tabular}{@{}lcccc@{}}
\toprule
Layers & Event $\uparrow$ & State $\uparrow$ & Tendency $\uparrow$ & FMD $\downarrow$ \\
\midrule
None & --- & --- & --- & 19.3 \\
T & --- & --- & 0.599 & 19.5 \\
S & --- & 0.315 & --- & 20.3 \\
E & 0.686 & --- & --- & 16.7 \\
T + S & --- & 0.314 & 0.593 & 20.2 \\
T + E & 0.680 & --- & 0.613 & 17.1 \\
S + E & 0.742 & 0.333 & --- & 17.1 \\
T + S + E & 0.746 & 0.334 & 0.609 & 17.3 \\
\bottomrule
\end{tabular}
\caption{Stacking the three layers. T = tendency, S = state, E = event, taken from the real standard notes of each clip. Execution rates are measured against what is actually specified under each condition; ``---'' means the layer is not specified. FMD is the absolute velocity FMD over all frames, including the constrained ones, and cannot be compared directly with Tab.~1.} %
\label{tab:supp_layers}
\end{table}

\begin{table}[tb]
\centering
\footnotesize
\setlength{\tabcolsep}{2pt}
\begin{tabular}{@{}lcccccc@{}}
\toprule
Density & Constr. & Event & State & Tend. & FMD & F1 (real) \\
\midrule
Empty & 0 & --- & --- & --- & 19.3 & 0.17 \\
$K=0$ & 0 & --- & 0.320 & 0.593 & 20.2 & 0.19 \\
$K=6$ & 0.021 & 0.714 & 0.340 & 0.603 & 18.0 & 0.26 \\
$K=15$ & 0.027 & 0.726 & 0.334 & 0.607 & 17.9 & 0.27 \\
$K=30$ (standard) & 0.044 & 0.746 & 0.334 & 0.609 & 17.3 & 0.32 \\
$K=60$ & 0.072 & 0.750 & 0.357 & 0.632 & 15.6 & 0.42 \\
Full & 0.128 & 0.698 & 0.422 & 0.688 & 11.1 & 0.66 \\
\bottomrule
\end{tabular}
\caption{Density sweep. Constr.: fraction of constrained entries. Event, state and tendency execution are against
what each condition specifies; F1 (real): event F1 against the real full notes; FMD: all-frame, absolute (as in
\cref{tab:supp_layers}). $K=0$ keeps the tendencies and one state segment per minute.}
\label{tab:supp_density}
\end{table}

\subsection{Emotion}
\label{sec:supp_emotion}

The 184 clips consist of 8 clips for each of the 8 emotions of RAVDESS~\cite{livingstone2018ravdess} and 20 for each of the 6 emotions of CREMA-D~\cite{cao2014cremad};
177 of them are among the 299 clips of Tab.~2. The language model authors notes following a direction of the form
``say this line with a certain emotion''. The classifier (one per dataset) maps 378 per-channel statistics through feature standardization and a multinomial logistic regression trained on real training clips ($C=0.01$ by speaker-grouped
5-fold cross-validation (CV); CV accuracy 0.567 / 0.623, real validation clips 0.703 / 0.642 for RAVDESS / CREMA-D; chance 0.125 /
0.167). The standardization is refitted per condition: with the training-set standardization, all generated results
are close to chance (0.14--0.28), while real clips stay at 0.65--0.72.
The per-condition refit uses no labels, is applied to real clips in the same way, and is the primary measure of
our scoring protocol. We attribute the near-chance results to an overall shift of the feature distribution of all
generated motion relative to real motion (the per-channel statistics differ in overall level), which every format
shares: under the training-set standardization the classifier mainly responds to this shared shift, so every
generated condition is near chance and the formats cannot be told apart. With per-condition standardization, the
classifier compares the clips within one condition, and every format is treated the same way. The intensity slope
instead uses one standardization per format, fitted on its three intensity conditions together, because a
per-condition refit would remove the change in overall amplitude that intensity itself causes. \Cref{tab:supp_emotion} gives all conditions.
With Opus, LDN exceeds TL by $+0.120$ $[0.052, 0.182]$ and TC16 by $+0.156$ $[0.104, 0.219]$ on RAVDESS. With Haiku,
about 20\% of the notes are invalid, LDN has no accuracy advantage, and its intensity slope is still above that of TL.

\begin{table}[tb]
\centering
\footnotesize
\setlength{\tabcolsep}{2pt}
\resizebox{\linewidth}{!}{%
\begin{tabular}{@{}llccc@{}}
\toprule
 & & \multicolumn{2}{c}{Accuracy (R / C)} & Slope (R / C) \\
\cmidrule(lr){3-4}\cmidrule(l){5-5}
Author & Format & emotion & neutral & a little $\to$ very \\
\midrule
Opus & LDN & 0.724 / 0.592 & 0.536 / 0.458 & 0.126 / 0.151 \\
 & TL & 0.604 / 0.556 & 0.531 / 0.489 & 0.045 / 0.043 \\
 & TC16 & 0.568 / 0.525 & 0.516 / 0.486 & 0.031 / 0.050 \\
Haiku & LDN & 0.604 / 0.506 & 0.557 / 0.447 & 0.091 / 0.047 \\
 & TL & 0.604 / 0.567 & 0.552 / 0.489 & 0.023 / $-$0.007 \\
 & TC16 & 0.547 / 0.550 & 0.484 / 0.481 & 0.028 / 0.035 \\
\midrule
\multicolumn{2}{@{}l}{Real motion} & 0.703 / 0.642 & & \\
\multicolumn{2}{@{}l}{Real standard notes, LDN} & 0.635 / 0.542 & & \\
\multicolumn{2}{@{}l}{Empty notes} & 0.516 / 0.475 & & \\
\bottomrule
\end{tabular}}
\caption{Emotion (R: RAVDESS, C: CREMA-D). Emotion: notes requesting the target emotion; neutral: the same author
without an emotion request. Slope: change in the classifier's probability of the target emotion from ``a little'' to
``very''.}
\label{tab:supp_emotion}
\end{table}

\subsection{Language-Model Authors}

The strict validity rate of director notes authored by Claude Opus 5.5 is 96\%, and after a fixed set of format repairs it is 99\%; after the same repairs, Haiku reaches 94\% and Gemini 3.1 Pro 98\%. The validity rate of blendshape keyframes is also high after the repairs, but their strict validity rate is much lower (Tab.~4). %
Haiku's errors are mostly invented field formats (\eg, giving a head direction as \texttt{x\_down}), which the fixed
rules repair. The keyframe formats are interpolated linearly between keyframes, and the face barely moves. Model
versions: claude-opus-5-5, claude-haiku-5-5 and gemini-3.1-pro-preview (thinking level low), otherwise default
settings. Paired permutation tests of the relative FMD in Tab.~4 (LDN vs.\ each format): Opus TL $p=0.04$, TC16
0.84, BS and AU $<0.001$; Haiku TL 0.002, TC16 0.27, BS and AU $<0.001$. The difference to TL changes direction
across datasets (LDN better on CREMA-D, worse on TalkVid).

\Cref{tab:supp_gemini} compares the three authors on the clips that Gemini authored. On these clips, Gemini's
validity and relative FMD fall within or above the range of Opus and Haiku, so the conclusions do not depend on one
vendor. On this subset the keyframe formats have a relative FMD of only 1.15--1.45, far below 3.03--3.52 on all 299
clips, for Opus and Haiku as well; the table is therefore only a comparison between authors. Gemini permutation
tests: TL $p=0.81$, TC16 0.34, BS 0.001, AU 0.14.

\begin{table}[tb]
\centering
\footnotesize
\setlength{\tabcolsep}{1.5pt}
\resizebox{\linewidth}{!}{%
\begin{tabular}{@{}lcccccc@{}}
\toprule
 & \multicolumn{3}{c}{Validity (repaired / strict)} & \multicolumn{3}{c}{Relative FMD} \\
\cmidrule(lr){2-4}\cmidrule(l){5-7}
Format (clips) & Opus & Haiku & Gemini & Op. & Ha. & Ge. \\
\midrule
LDN (56 / 54) & 1.00 / 0.98 & 0.93 / 0.59 & 0.98 / 0.91 & 1 & 1 & 1 \\
TL (59 / 54) & 1.00 / 0.95 & 0.98 / 0.92 & 1.00 / 1.00 & 0.98 & 1.00 & 1.01 \\
TC16 (59 / 54) & 1.00 / 0.98 & 1.00 / 1.00 & 1.00 / 1.00 & 0.99 & 1.03 & 1.03 \\
BS (30 / 27) & 1.00 / 0.77 & 1.00 / 0.37 & 1.00 / 0.77 & 1.18 & 1.22 & 1.45 \\
AU (30 / 27) & 1.00 / 0.60 & 1.00 / 0.47 & 1.00 / 0.90 & 1.15 & 1.16 & 1.15 \\
\bottomrule
\end{tabular}}
\caption{Three authors on the clips Gemini authored (clips for validity / for FMD; FMD only on clips where all
three produced valid notes). Relative FMD: all-frame, LDN of the same author = 1.}
\label{tab:supp_gemini}
\end{table}

\setcounter{topnumber}{2}
\subsection{Vision--Language Reader}
\label{sec:supp_vlm}

Our reader first re-estimates the 63-d curves from every frame with the same tracking pipeline as the training data (MediaPipe~\cite{lugaresi2019mediapipe}), and the encoder then converts them into director notes; the whole process is deterministic. The
vision--language model (VLM), Gemini (gemini-3.1-pro-preview, temperature 0), watches the same video and receives the
script, word timing and the same LDN specification, and authors director notes directly; its notes go through the
same repair and validation as the language-model authors, invalid notes count as empty, and 23 of 24 are valid.
Generation uses a single seed. The 24 clips are 8 per dataset, rendered with the ICT avatar~\cite{li2020ict} at 24 fps and
$512\times512$. %
\Cref{tab:supp_vlm} adds two references. Rendering real motion and recovering it with tracking + encoder gives 71\% of the events,
while reading real parameters directly recovers almost all (Sec.~4.2), so rendering and tracking alone lose about
30\% of the events; generated motion read from video (59\%) is at the same level, so the low score of the
vision--language model is not because the motion is invisible. Reading the generated parameters directly, as in the
cycle-consistency evaluation, recovers 78\%. Per dataset, tracking + encoder / vision--language model recover 0.59 /
0.27 (TalkVid), 0.50 / 0.12 (RAVDESS) and 0.69 / 0.08 (CREMA-D). Intervals: parameters vs.\ the vision--language model
$+0.61$ $[0.44, 0.77]$; tracking + encoder vs.\ the vision--language model $+0.41$ $[0.24, 0.59]$.

\begin{table}[tb]
\centering
\footnotesize
\setlength{\tabcolsep}{1.8pt}
\resizebox{\linewidth}{!}{%
\begin{tabular}{@{}llccccc@{}}
\toprule
Reader & Input & Event & State & Mouth t. & Smile & Head r. \\
\midrule
Track.\ + enc. & video of generated & 0.588 & 0.322 & 7/14 & 4/4 & 5/11 \\
VLM & video of generated & 0.176 & 0.034 & 0/14 & 0/4 & 1/11 \\
\midrule
\textit{Track.\ + enc.} & \textit{video of real} & \textit{0.706} & \textit{0.407} & \textit{8/14} & \textit{4/4} & \textit{9/11} \\
\textit{Encoder} & \textit{generated params} & \textit{0.784} & \textit{0.424} & \textit{13/14} & \textit{4/4} & \textit{6/11} \\
\bottomrule
\end{tabular}}
\caption{Reading director notes from silent renders (24 validation clips, 51 specified events). Event: recall; State:
state execution; last three columns: recovered events per family. Italic rows: references.}
\label{tab:supp_vlm}
\end{table}

\subsection{Full Ablation}
\label{sec:supp_ablation}

\begin{table*}[t]
\centering
\footnotesize
\setlength{\tabcolsep}{4pt}
\begin{tabular}{@{}llcccccccc@{}}
\toprule
Group & Configuration & FMD$\downarrow$ & W1$\downarrow$ & Seam$\downarrow$ & Event$\uparrow$ & State$\uparrow$ & Head jitter & Blink err.$\downarrow$ & Mouth$\uparrow$ \\
\midrule
Conditioning (50k iter.) & Script only & 1.17 & 1.16 & 2.39 & 0.172 & 0.128 & 3.18 & 0.204 & 0.232 \\
 & LDN only & 0.97 & 0.98 & 1.00 & 0.827 & 0.314 & 2.55 & 0.230 & 0.115 \\
 & LDN + script \checkmark & 1.00 & 1.00 & 1.00 & 0.830 & 0.324 & 2.64 & 0.263 & 0.299 \\
\midrule
Training window (15k iter.) & 3.2\,s \checkmark & 1.00 & 1.00 & 1.00 & 0.818 & 0.299 & 3.50 & 0.325 & 0.186 \\
 & 6.4\,s & 1.01 & 1.13 & 1.17 & 0.814 & 0.305 & 3.84 & 0.314 & 0.183 \\
 & Whole clip & 1.04 & 1.15 & 1.19 & 0.821 & 0.313 & 3.77 & 0.323 & 0.187 \\
\midrule
Training variants (15k iter.) & Base & 1.03 & 0.98 & 0.97 & 0.818 & 0.299 & 3.50 & 0.325 & 0.186 \\
 & Coarse base + residual & 1.17 & 1.01 & 1.07 & 0.857 & 0.291 & 2.82 & 0.362 & 0.157 \\
 & Script perturbation \checkmark & 1.00 & 1.00 & 1.00 & 0.819 & 0.305 & 3.50 & 0.314 & 0.151 \\
 & Condition dropout & 0.99 & 1.12 & 1.21 & 0.804 & 0.292 & 3.89 & 0.315 & 0.187 \\
 & Jump penalty & 1.40 & 1.00 & 0.56 & 0.814 & 0.282 & 3.21 & 0.434 & 0.154 \\
 & Left--right mirroring & 1.06 & 1.16 & 1.12 & 0.816 & 0.295 & 3.86 & 0.306 & 0.172 \\
 & No visemes & 1.03 & 0.98 & 0.95 & 0.816 & 0.300 & 3.48 & 0.309 & 0.132 \\
\midrule
Blink masking $p$ & 0 & 1.08 & 1.02 & 0.99 & 0.745 & 0.324 & 2.59 & 0.248 & 0.222 \\
 & 0.3 & 1.04 & 1.01 & 1.01 & 0.749 & 0.325 & 2.59 & 0.210 & 0.222 \\
 & 0.6 & 1.01 & 1.00 & 1.00 & 0.744 & 0.327 & 2.60 & 0.194 & 0.222 \\
 & 0.9 \checkmark & 1.00 & 1.00 & 1.00 & 0.745 & 0.326 & 2.60 & 0.175 & 0.222 \\
\midrule
Refinement & None \checkmark & 1.00 & 1.00 & 1.00 & 0.745 & 0.326 & 2.60 & 0.175 & 0.222 \\
 & Refinement net 1 & 7.48 & 2.96 & 8.63 & 0.532 & 0.244 & 0.26 & 0.946 & 0.238 \\
 & Refinement net 2 & 7.19 & 2.94 & 3.34 & 0.746 & 0.247 & 0.32 & 0.947 & 0.238 \\
\midrule
DDIM steps (no smoothing) & 25 \checkmark & 1.00 & 1.00 & 1.00 & 0.745 & 0.326 & 2.60 & 0.175 & 0.222 \\
 & 50 & 0.83 & 1.25 & 1.33 & 0.741 & 0.329 & 3.10 & 0.135 & 0.216 \\
 & 100 & 0.75 & 1.45 & 1.53 & 0.739 & 0.331 & 3.42 & 0.117 & 0.213 \\
\midrule
Head smoothing $\sigma_h$ & 25 steps, none & 0.94 & 1.06 & 3.41 & 0.745 & 0.326 & 2.60 & 0.175 & 0.222 \\
 & 25 steps, 0.5 & 0.97 & 0.99 & 1.93 & 0.745 & 0.326 & 1.75 & 0.175 & 0.222 \\
 & 25 steps, 0.75 \checkmark & 1.00 & 1.00 & 1.00 & 0.743 & 0.326 & 0.92 & 0.175 & 0.222 \\
 & 25 steps, 1.0 & 1.04 & 1.03 & 0.98 & 0.741 & 0.326 & 0.62 & 0.175 & 0.222 \\
 & 100 steps, 0.75 & 0.75 & 1.25 & 1.95 & 0.738 & 0.331 & 1.16 & 0.117 & 0.213 \\
\bottomrule
\end{tabular}
\caption{All generator ablations (validation set, standard notes). \checkmark: the configuration selected in each group;
FMD, W1 and seam are normalized by it. Head jitter: relative to real motion (closer to 1
is better). Blink err.: relative error of the spontaneous eyeblink frequency on frames with no specified blink. Mouth:
Pearson correlation of the generated and real jawOpen curves (per clip and sample, averaged per dataset, macro over
datasets). The event recall of the selected smoothing row (0.743) is without the final projection; Tab.~1 (0.744)
includes it.}
\label{tab:supp_ablation}
\end{table*}

\Cref{tab:supp_ablation} lists all generator ablations.
Coarse base + residual: the network predicts only the residual over a coarse base, the interpolation of the
decoded notes (as in R1). Jump penalty: an extra loss on the first and second time differences of the composite curve
(decoded values on the constrained entries, prediction elsewhere) against the ground truth ($\lambda=1$, weight 4
within 3 frames of a mask boundary). Refinement net 1: a separate small network applied after sampling that predicts a
residual to restore detail, trained on cross-fitted generator outputs; refinement net 2 is refinement net 1 followed
by the projection again. Head jitter: the median absolute second difference of the generated head rotation, divided
by that of real motion. Each group changes one choice and uses the default configuration at the time it was run, so values are not compared across groups. FMD, W1 and seam are normalized within each group by the selected
configuration (\checkmark); event recall and state execution are absolute. The conditioning, window and training-variant
groups use an earlier event template, so their event recall (about 0.82) is not comparable with the later groups
(about 0.745). The training-variant group uses a single seed except the base setting. Two choices did not follow the
pre-set selection rule, which looks only at naturalness: LDN was kept because without it event recall is only 0.17,
and condition dropout passed the rule but was not adopted because W1 and seam worsen. The selected script
perturbation lowers the mouth correlation from 0.186 to 0.151, which the rule did not consider; removing visemes
lowers it to 0.132. The two refinement networks differ only on the 32 non-mouth channels, so their mouth correlation is
identical. The conditioning-group value 0.299 and the final model's 0.222 come from different settings.

\subsection{Mouth}
\label{sec:supp_mouth}

Without audio, mouth motion is generated from word timing and visemes. The difference in mouth opening between frames inside
and outside words is 0.85 of that in real speech, but the per-frame jawOpen correlation with the real curve is weak
(0.22; both are macro averages over the three datasets). Per dataset (TalkVid / RAVDESS / CREMA-D), the final model reaches 0.054 / 0.349 / 0.262; in the
conditioning ablation, LDN + script gives 0.079 / 0.448 / 0.369, script only 0.043 / 0.381 / 0.271 and LDN only
0.027 / 0.169 / 0.150. TalkVid stays at or below 0.08 for every configuration.

\paragraph{Combining with an audio-driven lip model.} The 14 speech channels (\cref{tab:supp_families}) can be taken
from Audio2Face-3D~\cite{nvidia2025audio2face} driven by Kokoro TTS~\cite{hexgrad2025kokoro} of the same words, stretched to the word timing (the A2F row of Tab.~1), while
all other channels come from our generator; the two sets do not overlap and the speech channels are never
constrained. Event recall (0.744) and state execution (0.326) are unchanged, with identical per-dataset event counts
for smile and mouth tension; tendency execution drops from 0.615 to 0.573, and the relative FMD is 1.02. The
correlation with the real jawOpen drops from 0.22 to 0.02, as expected, since TTS differs from the real speech. We
also ran SyncNet~\cite{chung2016syncnet} on LAM renders (10 clips): LSE-C~\cite{prajwal2020wav2lip} is 0.31 / 0.50 / 0.85 for ours / combined / A2F only, against 0.32
/ 0.40 / 0.67 with mismatched audio. The differences are small, so SyncNet separates these renders only weakly, and
we do not draw conclusions from it. Videos 1--2 of the project page show the combination.

\subsection{Agreement with Human Annotation}

The 24 validation clips are 12 TalkVid, 5 RAVDESS and 7 CREMA-D clips with 256 annotated events. One annotator marked
the approximate position and family of every short event on the raw video, without seeing the encoder output.
Because the marks are approximate, Sec.~4.2 matches an annotated and an encoded event of the same family one-to-one if
their centres lie within 0.5\,s (greedy by distance), which gives a micro-F1 of 0.53, a macro-F1 of 0.44 and 0.88 for
blinks.

\section{User Study Protocol}
\label{sec:supp_user}

The protocol was written before any result was seen; all later changes are listed below.

\paragraph{Stimuli.} All clips come from the TalkVid validation set and are rendered silent on LAM avatars with the
demo settings (\cref{sec:supp_render}). \emph{Naturalness}: 6-s windows (180 frames) of ours, R2, A2F+LDN and real
motion, all taken from the standard-notes row of Tab.~1; both sides of a trial use the same avatar. \emph{Intensity}:
on clips not used for naturalness, the notes keep only the clip's own tendencies plus one tested instruction,
specified as slight or strong (smile state 0.20 / 0.75, brow raise state 0.14 / 0.48, nod event 3.5$^\circ$ / 12$^\circ$),
each value inside its level's range. The two versions share the checkpoint and the noise and differ only in this
instruction. TL receives the same notes converted to its 45 labels; since it has no intensity, its slight and strong
inputs, and outputs, are identical, so 0.5 is expected. Two further trials specify only a very low or very high
eyeblink frequency. Only the action name, not the level, is shown above the videos. The stimuli were selected at random from a
larger rendered pool, at the level of whole units (all conditions of a unit together): 13
clip--avatar pairs for naturalness (ours against R2, A2F+LDN or real motion, counterbalanced over three groups) and
7 intensity trials. %

\paragraph{Procedure.} Each trial shows two clips side by side (position randomized); participants choose the one
that moves more naturally, or the one with the stronger action. Trial order is randomized within each part, with a
seed derived from the participant ID. Each participant answers 22 trials (13 naturalness, 7 intensity, 2 attention
checks), about 8 minutes. Participants are a convenience sample.

\paragraph{Analysis.} We report the share of trials choosing ours (naturalness) or the strong version (intensity),
with 95\% intervals from a participant bootstrap (10000 resamples) and a two-sided exact binomial test against 0.5.
Per family, the strong version is identified at 0.77 (smile), 0.90 (brow raise) and 0.87 (nod) with ours, and at
0.43, 0.47 and 0.63 with TL ($p\ge0.20$).

\paragraph{Changes after the protocol.} (1) After 3 submissions, the two attention checks were dropped as exclusion
criteria because both were unreliable (the strong-smile check used a soft condition that renders weakly; the still
face was read as ``more natural'' by some); they are reported only (18 of 30 passed). (2) After 9 submissions, a
speed rule was added: a submission whose median time per trial is below 3\,s is excluded (shorter than a third of a
clip); it was set after seeing its effect on one submission, which made ``ours vs.\ real'' more conservative (0.50
$\to$ 0.46). (3) Group balance was enforced by distributing one fixed-group file per group. (4) Repeated
submissions of the same person were removed, keeping the first. In total, 40 submissions
were received; 4 repeats and 6 fast submissions were excluded, leaving 30.

\section{Rendering, Avatars and Demo}
\label{sec:supp_render}

Rendering is for visualization only; all evaluation uses the parameters. %
Portraits are animated with LAM~\cite{he2025lam} (pretrained and frozen; LAM-20K weights, CC BY-NC 4.0), which
builds a Gaussian head from one image. Our blendshapes drive it through a ridge regression ($\alpha=0.1$, chosen by
speaker-grouped 5-fold CV) from 41 MediaPipe expression channels to FLAME~\cite{li2017flame} expression and jaw
parameters, fitted on 2858 paired TalkVid training frames from 378 speakers; the target is LAM's own single-image
FLAME tracker on the same frame, and the CV vertex-displacement $R^2$ is 0.65 (jaw 0.45). Renderer settings: region
gain forehead 2.5, eyes 2.5, cheeks 1.8; head rotation limited to 15$^\circ$. Videos and user-study stimuli use an
expression display gain of 1.3 around the clip median; the frames of Fig.~5b use a gain of 2.5 around the most
relaxed frame, applied identically to every column, with levels smile 0.12 / 0.43 / 1.0, brow raise 0.11 / 0.27 / 0.9
and nod 3 / 6.5 / 15$^\circ$. The generated curves are never changed. All avatar portraits were generated from text
descriptions with Gemini 3 Pro Image and depict no real person; \cref{fig:supp_notes} shows frames of a real TalkVid
training clip.

The project page, linked from the code repository, shows the videos of Fig.~5, notes authored by hand, notes authored by language models from a script,
and the combination with an audio-driven lip model. In the demos, the script timing comes from Kokoro TTS~\cite{hexgrad2025kokoro} with MMS forced
alignment~\cite{pratap2024mms}. The output of the LDN system, per-frame curves of 52 blendshape weights, head pose and gaze (63 values per frame), drives
blendshape avatars directly and portrait images through the pretrained and frozen LAM. %

\section{Limitations}
\label{sec:supp_limits}

States reach 42--94\% of the specified value, and when one tendency is swept from very low to very high, its order is
kept but its exact level is recovered only 23--30\% of the time. The encoder that provides the training labels also
provides the evaluation readings; it agrees with a human annotator on blinks and on the timing of head and gaze
motion, but marks more small brow and squint events (Sec.~4.2). In our evaluation the clip header (neutral values and
reference orientation) comes from the real clip; for a new character it has to be taken from a reference clip or set
by the user, as in our demos. The LDN system takes no audio, so mouth motion is generated from word timing and
visemes rather than synchronized to speech (\cref{sec:supp_mouth}). A single-layer format with the same information as LDN can only be obtained by flattening layered notes
(\cref{sec:supp_flat}). FMD is computed on
parameters, not rendered video.

\end{document}